\documentclass[runningheads]{llncs}
\usepackage[margin=2.5cm]{geometry}
\usepackage{graphicx}

\usepackage{tikz}
\usepackage{comment}
\usepackage{amsmath,amssymb} 
\usepackage{color}
\usepackage{dsfont}
\usepackage[utf8]{inputenc}
\usepackage{algorithm}
\usepackage{algpseudocode}
\usepackage{caption}
\usepackage{xspace}

\newcommand{\bx}{\mathbf{x}}
\newcommand{\bm}{\mathbf{m}}

\algrenewcommand\algorithmicindent{0.5em}%

\makeatletter
\DeclareRobustCommand\onedot{\futurelet\@let@token\@onedot}
\def\@onedot{\ifx\@let@token.\else.\null\fi\xspace}

\def\eg{\emph{e.g}\onedot} 
\def\ie{\emph{i.e}\onedot} 
\def\cf{\emph{c.f}\onedot}

\def\etal{\emph{et al}\onedot}
\makeatother

\begin{document}
\pagestyle{headings}
\mainmatter
\def\ECCVSubNumber{2901}  

\title{Improved Masked Image Generation with Token-Critic\\} 


\titlerunning{Improved Masked Image Generation with Token-Critic}
%
\author{José Lezama \and Huiwen Chang \and Lu Jiang \and Irfan Essa }
\institute{Google Research}

%
\authorrunning{}
%
\maketitle

\begin{abstract}
%
Non-autoregressive generative transformers recently demonstrated impressive  image generation performance, and orders of magnitude faster sampling than their autoregressive counterparts.  However, optimal parallel sampling from the true joint distribution of visual tokens remains an open challenge. In this paper we introduce Token-Critic, an auxiliary model to guide the sampling of a non-autoregressive generative transformer.  Given a masked-and-reconstructed real image, the Token-Critic model is trained to distinguish which visual tokens  belong to the original image and which were sampled by the generative transformer.  During  non-autoregressive iterative sampling, Token-Critic is used to select which tokens to accept and which to reject and resample. Coupled with Token-Critic, a state-of-the-art generative transformer significantly improves its performance, and outperforms recent diffusion models and GANs in terms of the trade-off between generated image quality and diversity, in the challenging class-conditional ImageNet generation.

\end{abstract}

\section{Introduction}

\begin{figure}[t]
\centering
\includegraphics[width=0.88\textwidth]{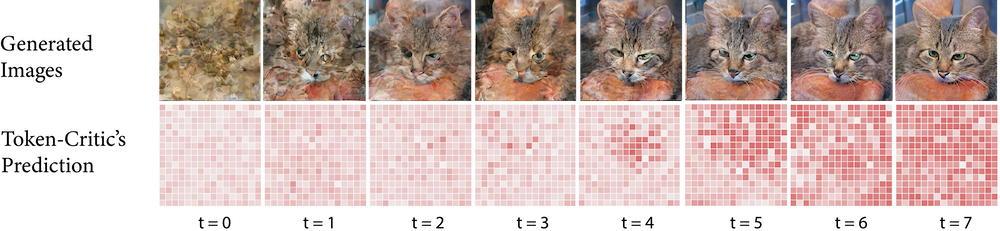}
\caption{Overview of the sampling procedure using Token-Critic. At each sampling iteration, Token-Critic predicts a high score for the tokens that are more likely sampled together under the joint distribution.
Tokens with lower score are masked and resampled at the next iteration.}
\label{fig:token-critic_sampling}
\end{figure}

Class-conditional image synthesis is a challenging task, requiring the generation of varied and semantically meaningful images with realistic details and few or none visual artifacts.
The field has seen impressive progress in the hand of mainly three techniques: large Generative Adversarial Networks (GANs) \cite{brock2018large}, diffusion models \cite{dhariwal2021diffusion,ho2022cascaded}, and transformer-based models over a vector-quantized (VQ) latent space \cite{esser2021taming,chang2022maskgit}.  Each of these techniques presents different advantages trading-off model size, computational cost of sampling, image quality and diversity. 

Building upon the transformers~\cite{vaswani2017attention} for the natural language generation tasks~\cite{gpt3}, generative vision transformers achieved impressive image generation performance. While early works applied an autoregressive transformer in the VQ latent space \cite{Oord17vqvae,esser2021taming}, recently the state-of-the-art on the common ImageNet benchmark was further advanced by a new model called MaskGIT \cite{chang2022maskgit} using mask-and-predict training inspired by BERT \cite{devlin2018bert} and non-autoregressive sampling adapted from neural machine translation \cite{ghazvininejad2019mask,kong2020incorporating}. 

To be more specific, during inference, MaskGIT \cite{chang2022maskgit} starts from a blank canvas with all the tokens masked out. In each step, it predicts all tokens in parallel but only keeps the ones with the highest prediction scores. The remaining tokens are masked out and will be re-predicted (resampled) in the next iteration until all tokens are generated with a few iterations of refinement. The non-autoregressive nature of MaskGIT allows orders-of-magnitude faster sampling, generating an image typically in 8-16 steps as opposed to hundreds of steps in autoregressive transformers \cite{esser2021taming} and diffusion models~\cite{dhariwal2021diffusion,ho2022cascaded}.

One of the central challenges of iterative non-autoregressive generation is knowing how many and which tokens to keep and which to resample at each sampling step. For instance, MaskGIT~\cite{chang2022maskgit} uses a predefined masking schedule and keeps the predicted tokens for which the model's prediction is more confident. However, this procedure presents three notable drawbacks. First, to select tokens to resample, it relies on the generator's predicted confidences which can be sensitive to modeling errors. Secondly, the decision to reject or accept is made independently for each token, which impedes capturing rich correlations between tokens. In addition, the sampling procedure is greedy and ``non-regrettable'', which does not allow to correct previously sampled tokens, even if they become less likely given the latest context.

In this work, we propose \emph{Token-Critic}, a second transformer that takes as input the output of the generative transformer (or generator for short). 
Intuitively, the Token-Critic is trained to recognize configurations of tokens likely under the real distribution, and those that were sampled from the generator. During the iterative sampling process, the scores predicted by Token-Critic are used to select which token predictions are kept, and which are masked and resampled in the next iteration (\cf Fig.~\ref{fig:token-critic_sampling}). 

With Token-Critic we tackle the three aforementioned limitations: 1) the masking of tokens is delegated to the Token-Critic model, trained to distinguish which tokens are unlikely under the true distribution. 2) Token-Critic looks at the entire set of sampled tokens collectively, thus is capable of capturing (spatial or semantic) correlations between tokens. 3) The proposed sampling procedure allows to correct previously sampled tokens during the iterative decoding. 

When using Token-Critic, the state-of-the-art non-autoregressive generative transformer MaskGIT \cite{chang2022maskgit} significantly improves its performance on ImageNet 256$\times$256 and 512$\times$512 class-conditional generation, while achieving a better trade-off between image quality and diversity. Furthermore, the gain obtained by using Token-Critic is complementary to the gain obtain by a pretrained ResNet classifier for rejection sampling. When coupled with classifier-based rejection sampling \cite{razavi2019generating}, Token-Critic parallels or surpasses the state-of-the-art continuous diffusion models with classifier guidance~\cite{dhariwal2021diffusion} in image synthesis quality while offering two orders of magnitude faster in generating images during inference.

\section{Background}
\subsection{Non-autoregressive Generative Image Transformer}
Generally, transformer-based models generate images in two stages~\cite{Ramesh21dalle,esser2021taming}. First, the image is quantized into a grid of discrete tokens by a Vector-Quantized (VQ) autoencoder built upon VAE~\cite{Oord17vqvae,razavi2019generating}, GAN~\cite{esser2021taming}, or vision transformer backbones~\cite{vim2021}, in which each token is represented as an integer index in a codebook. In the second stage, an autoregressive transformer decoder~\cite{chen2020imagegpt} is learned on the flattened token sequence to generate image tokens sequentially based on the previously generated result (\ie autoregressive decoding). In the end, the generated codes are mapped to pixel space using the decoder obtained from the first stage.

Non-autoregressive transformers~\cite{ghazvininejad2019mask,gu-kong-2021-fully,kong2020incorporating}, which were originally proposed for machine translation, are, very recently, extended to improve the second stage of autoregressive decoding~\cite{kong2021blt}. For example, MaskGIT \cite{chang2022maskgit} demonstrates highly-competitive fidelity and diversity of conditional image synthesis on the ImageNet benchmark as well as faster inference than the autoregressive transformer~\cite{esser2021taming} in addition to the diffusion models~\cite{dhariwal2021diffusion,nichol2021improved}. To be specific, MaskGIT is trained on the masked language modeling (MLM) proxy task proposed in BERT~\cite{devlin2018bert}. During inference, the model adopts a non-autoregressive decoding method to synthesize an image in a constant number of steps (typically 8-16 steps)~\cite{gu-kong-2021-fully}. 
Starting with all the tokens masked
out, in each inference step, MaskGIT predicts all tokens simultaneously in parallel and
only keeps the ones with the highest prediction scores. The remaining tokens
are masked out and will be re-predicted in the next iteration. The mask ratio
is made decreasing, according to a cosine function, until all tokens are
generated.
In the following, Sections~\ref{sec:challenges_training} and \ref{sec:challenges_sampling} describe limitations in the training and sampling of the MaskGIT model. Then, in Section~\ref{sec:method} we introduce the Token-Critic as a solution to mitigate these limitations.

\subsection{Challenges in Training Non-Autoregressive Transformers}\label{sec:challenges_training}
Ideally, one would like the masked generative transformer to learn the joint distribution of unobserved tokens $\bx = [x_1,\ldots x_N]$ given the observed tokens $\mathbf{o}$. Both $\bx$ and $\mathbf{o}$ are sequences of $N$ tokens where $N$ (\eg, $16 \times 16$) indicates the latent size of the VQ autoencoder obtained in the first stage. Each $x_j \in \mathcal{V}=\{1,\cdots,K\}$ is an integer token in the codebook of size $K$. Notice the element in $\textbf{o}$ can take the value of a special mask token, \ie $o_j \in \mathcal{V} \cup \{\texttt{[MASK]}\}$.
We shall refer to the true  distribution as $q\left(x_1,\ldots,x_N | \mathbf{o}\right)$.


Current non-autoregressive generative transformers \cite{chang2022maskgit,ghazvininejad2019mask} are trained to optimize the sum of \emph{the marginal} cross-entropies for each unobserved token:
\begin{equation}
     \mathcal{L}_i = - \sum_{j=1}^N \sum_{k=1}^K \tilde{q}(x_j=k | \mathbf{o}) \log p_\theta(x_j = k | \mathbf{o}),
\end{equation}
where $\tilde{q}$ represents an approximation to the true marginal given by considering one random real sample. 

A limitation is that optimizing over the marginals hinders capturing the richness of the underlying joint distribution of unobserved tokens. Essentially, this training scheme is equivalent to minimizing the Kullback-Leibler (KL) divergence between the data and model distributions, both approximated as fully factorizable distributions. 


\subsection{Challenges in Sampling from  Non-autoregressive Transformers}\label{sec:challenges_sampling}
During sampling, one is interested in sampling from the full joint distribution of unobserved tokens $q(x_1,\ldots x_N | \mathbf{o})$. However, even if the transformer representations are distributed, the  output for each token models its sampling distribution independently.  
More precisely, for a given unobserved token $x_t$, a value is sampled from $p_\theta\left(x_t| \mathbf{z}, \mathbf{o}\right) = p_\theta\left(x_t| \mathbf{z}\right)$, where $\mathbf{z}$ is the latent embedding 
visible by all output tokens (\ie the activations of the last attention layer). Sampling from the true distribution would require coordinating the values of all sampled tokens, which is not possible with the current architecture (unless the sampling is made deterministic, which harms the diversity of the generated images). Thus, non-autoregressive vision transformers still need to resort to iterative ancestor sampling. Typically, in each step of the sampling process, a growing subset of the tokens is accepted and the rest is rejected and resampled. 

Aiming at better approximating the true joint distribution, the question of how to select which sampled tokens to keep and which to resample is the main focus of this work.
We propose to do this using an auxiliary model that we term the Token-Critic. The Token-Critic is a second transformer trained to individually identify which tokens in a sampled vector-quantized image are plausible under the true joint distribution and which are not. During the iterative non-autoregressive sampling procedure, the Token-Critic is used in each iteration to reject the tokens that are less likely given the context.

\section{Method}\label{sec:method}

The goal of Token-Critic is to guide the iterative sampling process of
a non-autoregressive transformer-based generator.  Given the
tokenized image outputted by the generative transformer, Token-Critic is designed as a second transformer that provides a score for
each token, indicating whether the token is likely under the real
distribution, given its context.

In Section~\ref{sec:token_critic_training}, we first introduce the
procedure for training the Token-Critic and in
Section~\ref{sec:token_critic_sampling} we describe how it is used
during sampling. At all times we assume a pre-trained
non-autoregressive transformer generator is available. Finally, in
Section~\ref{sec:diffusion_connection} we explain the
role of the Token-Critic by drawing a connection to discrete diffusion processes.

\subsection{Training the Token-Critic}\label{sec:token_critic_training}

The training procedure for Token-Critic is  straightforward. Given a masked image and its corresponding completion by the generative non-autoregressive transformer, the Token-Critic is trained to distinguish which of the tokens in the resulting image were originally masked. 

More specifically, consider a real vector-quantized image $\bx_0$, a random binary mask $\bm_t$ and the resulting masked image $\bx_t = \bx_0 \odot \bm_t$. The subindex $t$ indicates the masking ratio, as will be detailed shortly. First, the generative transformer $G_\theta$, parameterized by $\theta$, is used to predict the masked tokens, namely sampling $\tilde{\bx}_0$ from $p_\theta(\hat{\bx}_0 | \bx_t, c)$, in which to condition on the class index $c$, we prepend a class token to the flattened set of visual tokens. The unmasked tokens in $\bx_t$ are copied into the output to form $\hat{\bx}_0 = \tilde{\bx}_0 \odot (1-\bm_t) + \bx_0 \odot \bm_t$. 

The Token-Critic transformer, parameterized by $\phi$, takes as input $\hat{\bx}_0$ and outputs a predicted binary mask for $\bm_t$. During training, the parameters $\phi$ are optimized to minimize the following objective:
\begin{align}
\mathcal{L}_i = \mathop{\mathbb{E}} \limits_{q(\bx_0, c)q(t)  q(\bm_t|t) p_\theta(\hat{\bx}_0|\bm_t \odot \bx_0, c)} \Big[ \sum_{j=1}^N BCE \Big(\bm_t^{(j)}, ~p_\phi(\bm_t^{(j)}|\hat{\bx}_0, c) \Big) \Big], \label{eq:token-critic_train}
%
\end{align}
where $q(\bx_0, c)$, $q(t)$, $q(\bm_t|t)$ are the distributions of real unmasked images, timesteps, and binary masks, respectively, and BCE denotes the binary cross-entropy loss. 
The sampling distribution $p_\theta(\hat{\bx}_0 |\bm_t \odot \bx_0)$ induced by the generator $G_\theta$ is held fixed during the training of the Token-Critic model.

The training algorithm is summarized as pseudocode in Algorithm~\ref{algo:training}. Notice that $\gamma(t) \in (0,1)$ in Step 4 is the cosine mask scheduling function. 
Given a uniform random number $t$ sampled from $q(t)=\mathcal{U}(0,1)$, the number of masked tokens in $\bm_t$ is computed as $r=\lceil N\cdot \gamma(t) \rceil$, where $N$ is the total number of tokens within an image. 



\subsection{Sampling with Token-Critic}\label{sec:token_critic_sampling}
During inference, we are interested in progressively replacing masked tokens with an actual code in the vocabulary. Starting from a fully masked image $\bx_T$ and the class condition $c$, we iteratively sample from $p(\bx_{t-1} | \bx_t, c)$, which  may be approximated by:

\begin{align}
p(\bx_{t-1} | \bx_t, c) &= \sum_{\bx_0} p_\phi(\bx_{t-1} | \bx_0, c) p_\theta(\bx_0 | \bx_t, c) \\ \label{eq:reverse_step}
                        &= \mathop{\mathbb{E}} \limits_{\bx_0 \sim p_\theta(\bx_0 | \bx_t, c)} \Big[ p_\phi(\bx_{t-1} | \bx_0, c) \Big]\\
                        &\approx p_\phi(\bx_{t-1}| \hat{\bx}_0, c), ~\hat{\bx}_0\sim p_\theta(\bx_0 | \bx_t,c ). \label{eq:one_sample}
\end{align}
In \eqref{eq:reverse_step}, we assume that $\bx_{t-1}$ is conditionally independent of $\bx_t$ given $\bx_0$. We will get back to this assumption shortly. In \eqref{eq:one_sample}, the expectation is empirically approximated using single sample Monte Carlo for $p_\theta(\bx_0 | \bx_t, c)$, which is obtained from the output of the generative transformer $G_\theta$. The next step is to sample from $p_\phi(\bx_{t-1}| \hat{\bx}_0, c)$. Recall that $\bx_{t-1}$ is a masked version of $\hat{\bx}_0$, rendering it solely determined by $\hat{\bx}_0$ and a mask $\bm_{t-1}$. Thus, we can sample $\bx_{t-1}$ in \eqref{eq:one_sample} using Token-Critic to predict a mask $\bm_{t-1}$ given $\hat{\bx}_0$. 

Note that the mask computation of MaskGIT \cite{chang2022maskgit} only relies on the prediction score $p_\theta(\bx_0 | \bx_t, c)$, in which tokens with the lowest predictions are masked. The mask sampling is independent for each token and moreover greedy which means previously unmasked tokens will be kept unmasked forever. In contrast, the proposed mask sampling is learned by the Token-Critic model $\phi$ to approximate sampling from the joint distribution by
taking into account the correlation among tokens. This notably improves the sampling leading to better generation quality. Secondly, Token-Critic makes generation regrettable, allowing to revoke prior decisions based on the most recent generation.


The sampling process is given as pseudocode in Algorithm~\ref{algo:sampling} and represented schematically in Figure~\ref{fig:token-critic_sampling}. The rate of masking in each step is given by the scheduling function $\gamma(t)$, with $t=T-1\ldots 0$, where higher values of $t$ correspond to more masking. After predicting $\bm_t$ in each step, we mask the $R=\lceil \gamma(t/T)\cdot N\rceil$ tokens with the lowest Token-Critic score. Following \cite{chang2022maskgit}, to introduce randomness in the first steps, we add a small ``selection noise'' $\mathbf{n}(t)$ to the Token-Critic scores before ranking them. This selection noise is annealed according to $\mathbf{n}(t) = K\cdot \mathbf{u} \cdot (t/T) $, where $K$ is a hyperparameter and $\mathbf{u} \in [-0.5,0.5]^N$ is a random uniform vector. Furthermore, the sampling temperature for each token is also annealed according to a linear schedule $T(t) = a\cdot(t/T)+b$.

Finally, we get back to the assumption in \eqref{eq:reverse_step} that $\bx_{t-1}$ is made independent of $\bx_t$ given $\bx_0$. This assumption can be dropped by simply adapting the Token-Critic's input by concatenating the previous mask $\bm_t$ to $\hat{\bx}_0$. However, in practice, we find this does not yield a better result. In fact, by ignoring the previous mask, Token-Critic has the ability to correct previously sampled tokens that are no longer as likely given the latest context, which addresses the greedy mask selection in the MaskGIT model \cite{chang2022maskgit}.

\begin{figure}[t]
\centering
\includegraphics[width=1.0\textwidth]{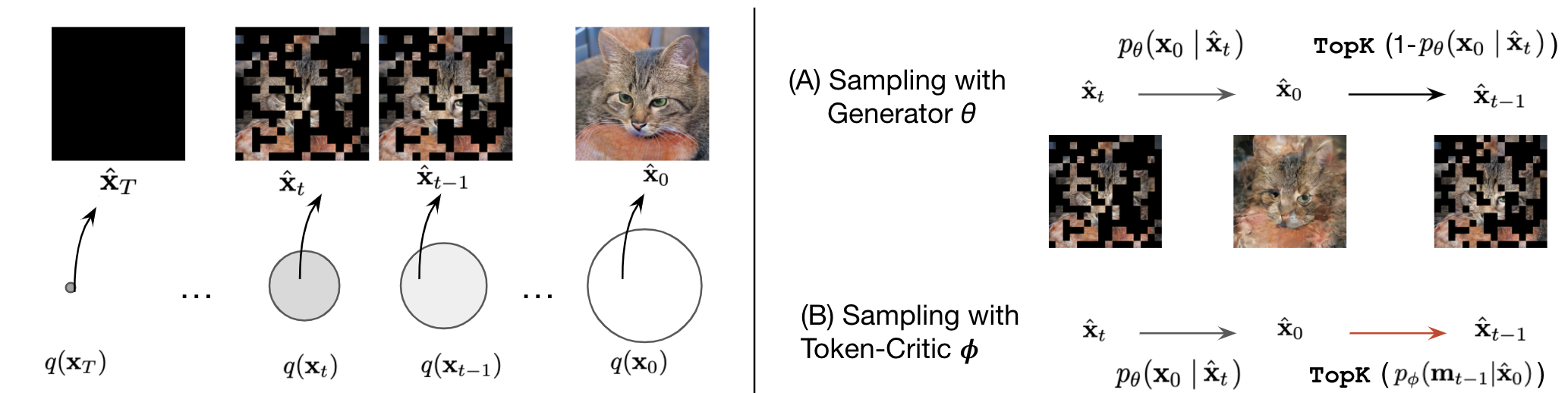}
\caption{Token-Critic in the lens of a discrete stochastic process that gradually masks a real image $\bx_0$ from $t=0\ldots T$, where $\bx_T$ is fully masked. The circles represent the distributions of real vector-quantized images under different masking rates. During the reverse process, a masked image estimate $\hat{\bx}_t$ is refined by first using the generator to predict a clean image $\hat{\bx}_0$, and then predicting the mask for the next timestep. While prior works \cite{chang2022maskgit} use the confidence of the generator $G_\theta$ (A), we use the predictions of Token-Critic (B) to select which tokens to mask. 
}
\label{fig:diffusion-diagram}
\end{figure}

\begin{table}[H]
    \centering
    \begin{tabular}{cc}
\begin{minipage}{0.49\textwidth}
\begin{algorithm}[H]
    \caption{Token-Critic Training}\label{algo:training}
     \noindent\textbf{Input:} Pre-trained generator $G_\theta$, scheduling function $\gamma(t)$, learning rate $\eta$ \\
     \noindent\textbf{Output:} Token-Critic parameters $\phi$  
    \begin{algorithmic}[1]
            \Repeat
        \State $\bx_0, c \gets \text{i.i.d. sampled VQ image}$
        \State $t \sim \mathcal{U}_{(0,1)}$
        \State $\bm_t \gets \text{random mask(} \lceil \gamma(t)\cdot N \rceil \text{)}$
        \State $\bx_t \gets \bx_0 \odot \bm_t$
        \State $\hat{\bx}_0 \gets G_\theta(\bx_t, c)$
        \State $\phi \gets \phi - \eta \nabla_\phi BCE(\bm_t, p_\phi(\bm_t | \hat{\bx}_0,c))$
                \Until $\text{convergence}$
    \end{algorithmic}
\end{algorithm}
\end{minipage}
&
\raisebox{1.1em}{
\begin{minipage}{0.49\textwidth}
\begin{algorithm}[H]
\caption{Token-Critic Sampling}\label{algo:sampling}
    \begin{algorithmic}[1]
    \State $\bx_T \gets [\texttt{[MASK]}]_{N}$
    \For{$t=T \ldots 1$}{}
    \State $k = \lceil \gamma((t-1)/T)\cdot N \rceil$
    \State $\hat{\bx}_0 = G_\theta(\bx_t, c)$
    \State $\{p_i\}_{i=1\ldots, N} \gets p_\phi(\bm_{t-1}^{(i)} | \hat{\bx}_0, c) + n(t)$
    \State $\tau \gets \text{ rank}_k(\{p_i\})$
    \State $\{\bm_{t-1}^{(i)}\} \gets 1 ~\text{if}~p_i > \tau,~ \text{else} ~0$
    \State $\bx_{t-1} = \hat{\bx}_0 \odot \bm_{t-1}$
    \EndFor
    \end{algorithmic}
\end{algorithm}
\end{minipage}
}
    \end{tabular}
    \label{tab:my_label}
\end{table}

\subsection{Relation to Discrete Diffusion Processes}\label{sec:diffusion_connection}

The role of Token-Critic can also be understood under the perspective of discrete diffusion processes \cite{austin2021structured,esser2021imagebart,gu2021vector,hoogeboom2021argmax}, where it is assumed that there exists a  stochastic process that gradually destroys information by masking. In this setting, the reverse process aims to progressively replace masked tokens with elements from the VQ codebook following the true distribution. In our case, this is what the generator transformer $G_\theta$ does in each step of the sampling procedure. Ideally, each intermediate result should lie within the  distribution of partially masked real images, since this is the distribution used to train $G_\theta$. The role of Token-Critic is to guide the intermediate samples towards these regions.

Figure~\ref{fig:diffusion-diagram} represents a schematic representation of the reverse sampling process.  Given a current estimate of a masked image $\hat{\bx}_t$, we use the generator to produce an estimate clean image $\hat{\bx}_0$. Note that due to the aforementioned modeling limitations, this estimate typically falls far from the distribution of real images. Token-Critic is then used to predict a less corrupted image $\hat{\bx}_{t-1}$ from $\hat{\bx}_0$. Since it was trained to distinguish incompatible tokens, the improved prediction is achieved by masking the least ``plausible-looking'' tokens. 

In the diffusion processes literature, a similar sampling strategy relying on an estimate of the clean image was used in \cite{song2020denoising} for the continuous case and \cite{hoogeboom2021argmax,austin2021structured,gu2021vector} for the discrete case. The difference in our approach is that we implicitly use a learned forward model instead of a fixed one obtained beforehand (\eg, the Gaussian prior). On the other hand, previous discrete diffusion models for image generation \cite{austin2021structured,esser2021imagebart,gu2021vector,hoogeboom2021argmax} typically assume a stochastic process that is independent for each token, and give a fixed form Markov chain that defines the probabilities of each token being masked, converted randomly or staying the same. Even under the independence assumption, if the number of token categories is large, the computation of the $n$-step Markov transition matrix required to obtain the posterior can be impractical. These design differences in part explain the diffusion models' low-efficiency when synthesizing high-resolution images. Instead, Token-Critic trades-off the analytical interpretability and tractability of these assumptions for a more efficient, learned forward process $p_\phi(\bx_t | \hat{\bx}_0)$. 

Finally, we can motivate the training objective of Token-Critic from the KL divergence between the distributions of real partially masked images $q(\bx_{t})$, and the distribution of partially masked images obtained in the intermediate steps by the proposed sampling scheme $p_{\theta,\phi}(\bx_t)$. We refer to the appendix for the derivation.

\section{Experiments}

\begin{figure}[t]
\centering
\begin{tabular}{cc}
\includegraphics[height=0.35\textwidth]{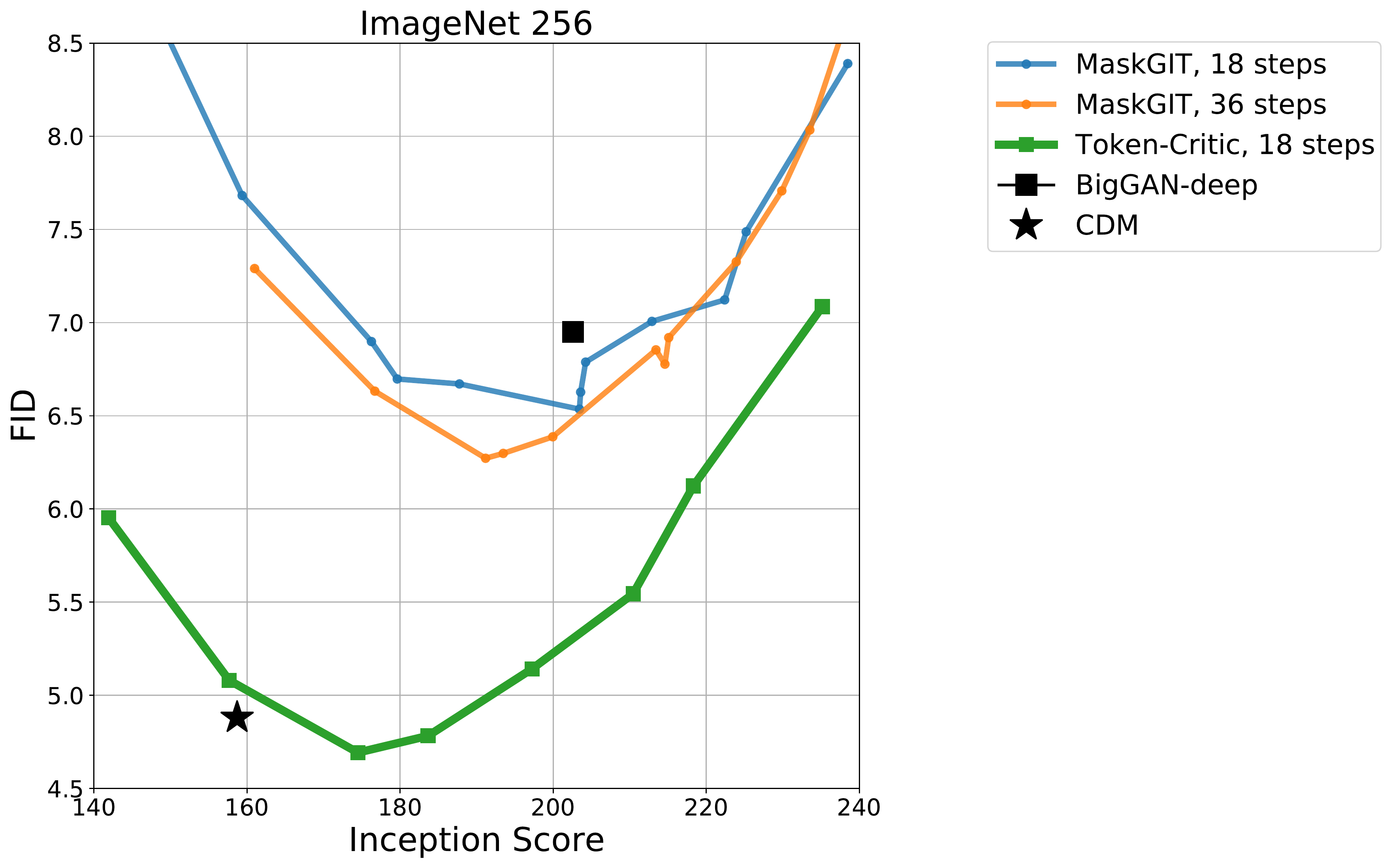}&
\includegraphics[height=0.35\textwidth]{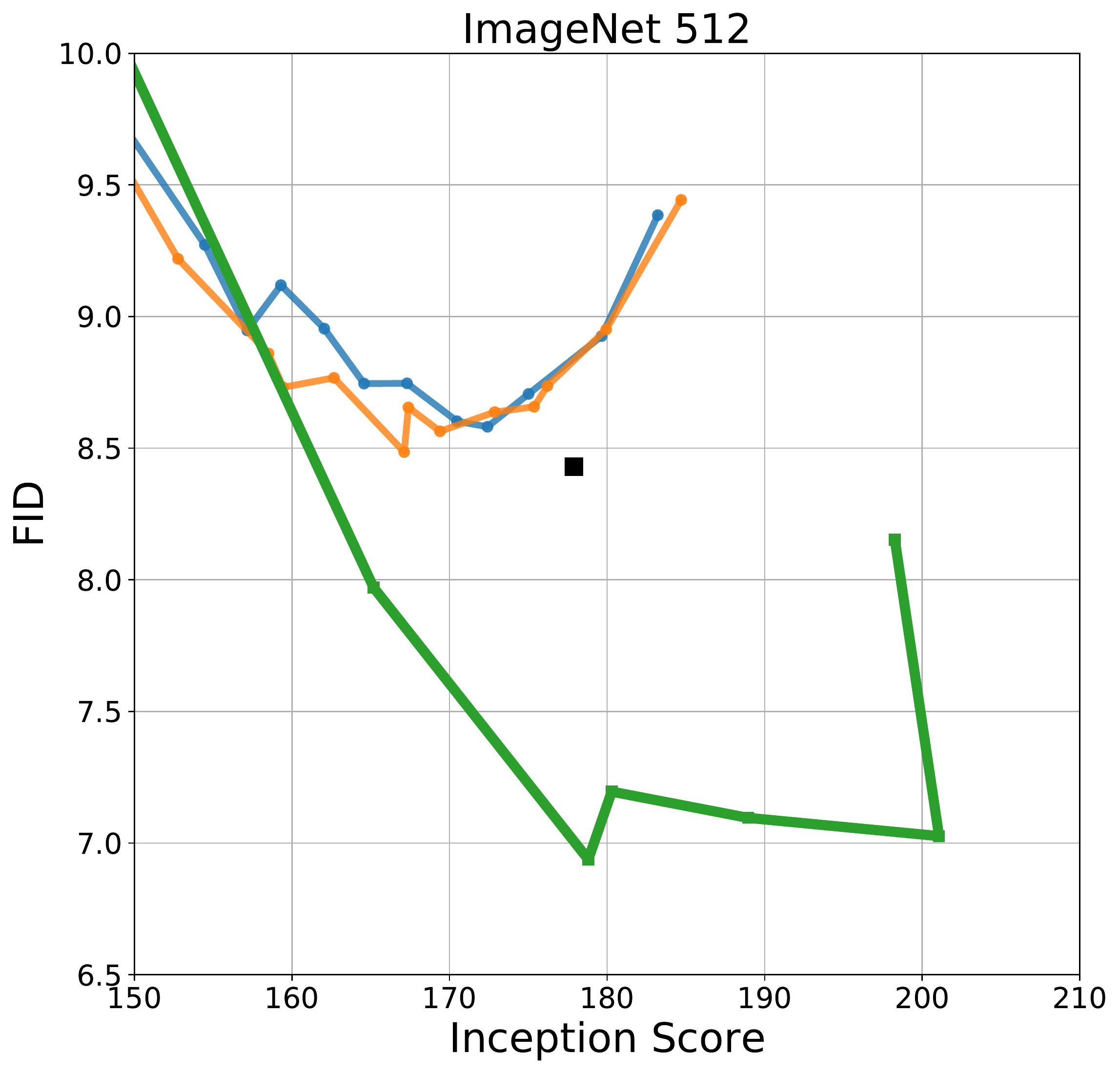}  
\end{tabular}
\caption{FID-vs-IS curves  on ImageNet 256x256 and 512x512 (bottom right is better). The trade-off between diversity and quality is traversed by varying the sampling temperature. Compared to the baseline \cite{chang2022maskgit}, sampling using Token-Critic produces a significant improvement in performance, and outperforms BigGAN-deep \cite{brock2018large} and CDM \cite{ho2022cascaded}, achieving a new state-of-the-art for methods that do not rely on an external classifier.  }
\label{fig:no_classifier_rejection}
\end{figure}

In this section we evaluate the proposed approach on class-conditional image generation tasks on ImageNet \cite{deng2009imagenet} 256$\times$256 and  512$\times$512. We compare over classical metrics to examine the trade-off between quality and diversity, notably FID \cite{heusel2017gans} vs. Inception Score \cite{salimans2016improved} and Precision vs. Recall \cite{kynkaanniemi2019improved}. We observed that the highly-competitive baseline is significantly improved when using Token-Critic, and that the proposed method obtains an advantageous quality-diversity trade-off, compared to state-of-the-art GANs and continuous diffusion models.

\subsection{Experimental Setup}
We use a pre-trained MaskGIT \cite{chang2022maskgit} model as the generator, and use the Token-Critic to guide the sampling, as described in Section~\ref{sec:token_critic_sampling}. We adopt the VQ encoder-decoder of \cite{esser2021taming} and \cite{chang2022maskgit}, with a codebook with 1024 tokens, trained at 256$\times$256 resolution in the same datasets. The VQ encoding compresses the image by a factor of 16, so that a 256$\times$256 (512$\times$512) image is represented as a grid of 16$\times$16 (32$\times$32) integers. The generator is a transformer with 24 layers and 16 heads. For the Token-Critic, we use a relatively smaller transformer with 20 layers and 12 heads, but otherwise of identical architecture. Both transformers use embeddings of dimension 768 and a hidden dimension of 3,072, learnable positional embedding \cite{vaswani2017attention}, LayerNorm \cite{ba2016layer}, and truncated normal initialization (stddev=0.02). The following
training hyperparameters were used for both MaskGIT and Token-Critic:  dropout
rate=0.1, Adam optimizer \cite{kingma2014adam} with $\beta_1=0.9$ and $\beta_2=0.96$.
We use RandomResizeAndCrop for data augmentation.
All
models are trained on 8$\times$8 TPU devices with a batch size of
256. The MaskGIT generators and the Token-Critic models were trained for 600 epochs. We use the same cosine schedule for the masking rate as in \cite{chang2022maskgit}, for both training and sampling. We use 18 steps when sampling with Token-Critic, as we found this gives the best results.

\begin{table}[t]
\centering
\begin{tabular}{ll@{\hspace{1em}}llll@{\hspace{1em}}|@{\hspace{1em}}llll}
 Model & steps & \multicolumn{4}{l}{~~ImageNet 256x256} &\multicolumn{4}{l}{~~ImageNet 512x512} \\[.5em]
                                &   & FID $\downarrow$ & IS $\uparrow$ &  Prec & Rec & FID $\downarrow$ & IS $\uparrow$ &  Prec & Rec\\
 \hline
 BigGAN-deep \cite{brock2018large} & 1&6.95       & 202.65 & \bf{0.86} & 0.24      & 8.43      & 177.9 & \bf{0.85} & 0.25      \\
 ADM \cite{dhariwal2021diffusion}  & 250 & 10.94      & 101.0      & 0.69      & \bf{0.63} & 23.24     & 58.06      & 0.73      & \bf{0.60} \\
 CDM \cite{ho2022cascaded}         & 100 & 4.88       & 158.7      & n/a       & n/a       & n/a       & n/a        & n/a       & n/a       \\
 MaskGIT \cite{chang2022maskgit}   & 18 & 6.56      & \textbf{203.6}      & 0.79      & 0.48    & 8.48   & 167.1      & 0.78       &  0.46    \\
 MaskGIT+Token-Critic            & 18(x2) &\bf{4.69} & 174.5      & 0.76      & 0.53      & \bf{6.80} & \bf{182.1}        & 0.73      & 0.50  
\end{tabular}
\caption{Comparison between methods that do not leverage an external classifier. We report the sampling configurations that obtain the best FID score for each method, and refer to Figure~\ref{fig:no_classifier_rejection} for the more comprehensive  trade-off between FID and Inception Score. All methods are evaluated on ImageNet training set. Results for \cite{brock2018large} are as reproduced by \cite{sauer2022stylegan}.}
\label{tab:no_classifier_rejection}
\end{table}

\subsection{Class-Conditional Image Synthesis}

\subsubsection{Quantitative Results}
We evaluate our method on class-conditional synthesis using ImageNet. Our main results are summarized in Table~\ref{tab:no_classifier_rejection}. For a more comprehensive quantitative comparison, we compare Inception Score vs. FID in Figure~\ref{fig:no_classifier_rejection}. These represent the trade-off between image quality, associated to Inception Score, and diversity or coverage, associated to FID. To traverse the quality-diversity trade-off for the baseline and Token-Critic, we modify the sampling temperature and selection noise parameter $K$ (Section~\ref{sec:token_critic_sampling}). Higher selection noise and temperature produce higher variability but lower quality.

We compare the proposed approach to the MaskGIT baseline which uses the generator's prediction confidence to select which tokens to reject in each step of the iterative sampling. To account for the fact that the proposed method requires two forward passes for each sampling step, in Figure~\ref{fig:no_classifier_rejection} we compare the proposed approach to the MaskGIT baseline for double the number of sampling steps. We also compare to state-of-the-art GAN architectures BigGAN-deep \cite{brock2018large}, and continuous Cascaded Diffusion Model (CDM) \cite{ho2022cascaded} and Ablated Diffusion Model (ADM) \cite{dhariwal2021diffusion} without external classifier guidance.

Of all the methods that do not rely on an external classifier, the proposed approach achieves the lowest FID, while providing an advantageous FID / Inception Score balance. Compared to the MaskGIT baseline, it achieves a significant improvement in terms of FID and Inception Score.

\subsubsection{Leveraging an  External Classifier}

\begin{table}[t]
\centering
\begin{tabular}{l@{\hspace{1em}}llll@{\hspace{1em}}|@{\hspace{1em}}llll}
 Model  & \multicolumn{4}{l}{~~ImageNet 256x256} &\multicolumn{4}{l}{~~ImageNet 512x512} \\[.5em]
                                   & FID $\downarrow$ & IS $\uparrow$ &  Prec & Rec & FID $\downarrow$ & IS $\uparrow$ &  Prec & Rec\\
 \hline
 ADM+Guid.  \cite{dhariwal2021diffusion}              & 4.59       & 186.7      & 0.82      & 0.52      & 7.72      & 172.7      & \bf{0.87} & 0.42 \\
 ADM+Guid.+Upsamp. \cite{dhariwal2021diffusion}     & 3.94       & 215.8      & \bf{0.83} & 0.53      & 3.85      & 221.7      & 0.84      & \bf{0.53} \\
 StyleGAN-XL ($\Psi=1.0$ ) \cite{sauer2022stylegan}     & \bf{3.26}  & 225.6      & 0.74      & 0.45      & \bf{3.58} & 219.8      & 0.73      & 0.43 \\
 MaskGIT \cite{chang2022maskgit} (a.r. 20\%)            & 4.70           & 266 &   0.80      & 0.48          & 5.13  & 250.7  &  0.79   & 0.47  \\
\cite{chang2022maskgit}+Token-Critic (a.r. 20\%)                            & 3.75       & \bf{287.0}      & 0.75      & \bf{0.55} & 4.03      & \textbf{305.2}      & 0.73      & 0.50  
\end{tabular}
\caption{Comparison between methods that use an external classifier during training or sampling. We report the sampling configurations that obtain the best FID score for each method. We refer to Figure~\ref{fig:classifier_rejection} to better appreciate the improvement obtained by Token-Critic in the trade-off between image quality and sample diversity. Sampling with Token-Critic and a classifier rejection scheme significantly improves the baseline and obtains the best Inception Score of all compared methods.}
\label{tab:classifier_rejection}
\end{table}

Classifier-guidance is a commonly adopted technique in diffusion models to improve class-conditional generation \cite{esser2021taming,razavi2019generating}, consisting on using the gradient of an external classifier to improve the class score of the sampled image, and thus render it more semantically meaningful. 
We show that the improvement obtained by leveraging an external pre-trained classifier is independent of the improvement brought by the Token-Critic. Moreover, the combination of both further improves the quantitative performance, achieving the highest reported Inception Score, and FID scores competitive with the most advanced GANs and Diffusion Models that use an external classifier.

\begin{figure}[t]
\centering
\begin{tabular}{cc}
    \includegraphics[height=0.35\textwidth]{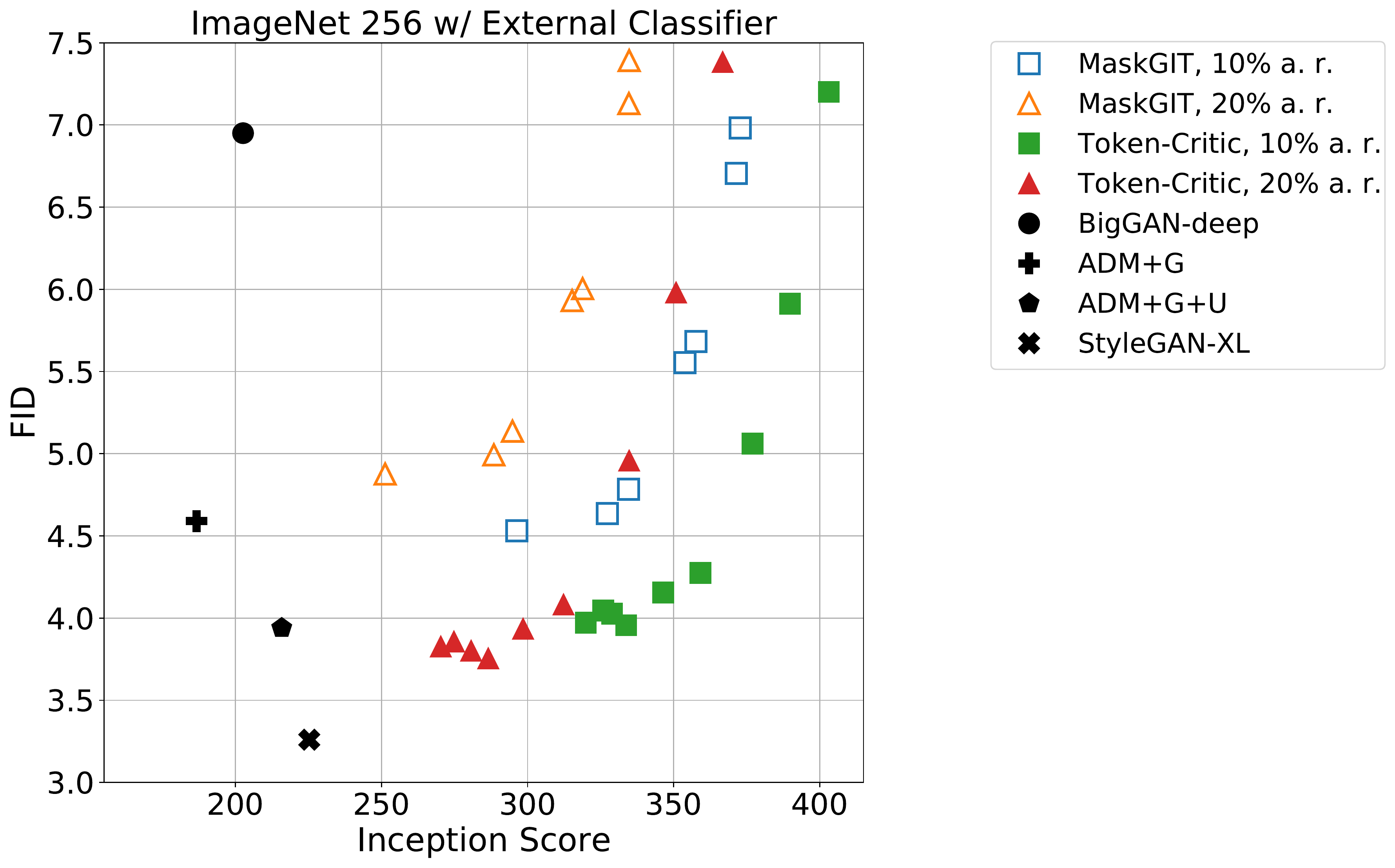}
 & \includegraphics[height=0.35\textwidth]{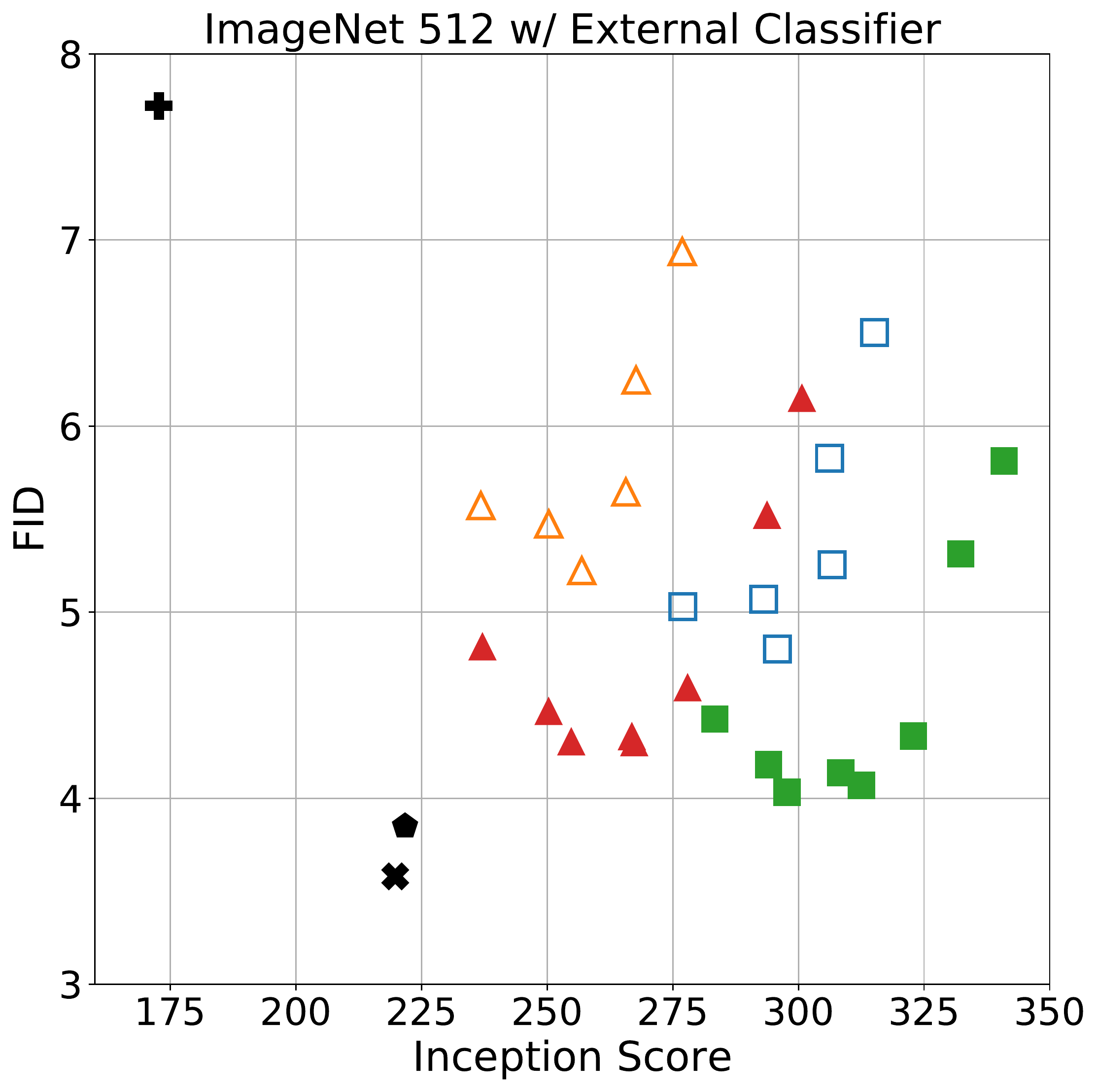}
\end{tabular}
\caption{FID-vs-Inception Score trade-off for methods that leverage an external classifier (bottom right is better). We use an external ResNet classifier for rejection sampling at acceptance rates 10\% and 20\%. Points in the graph indicate different sampling temperatures to balance quality and diversity, the remaining hyperparameters being equal. Token-Critic improves significantly upon  \cite{chang2022maskgit}, and also obtains a superior trade-off to diffusion models with upsampling \cite{dhariwal2021diffusion}. The trade-off is also comparable to the concurrent work StyleGAN-XL \cite{sauer2022stylegan}, which obtains the best FID but with much lower Inception Score. }
\label{fig:classifier_rejection}
\end{figure}

Since classifier guidance is not directly transferable to the VQ latent space, here we adopt a classifier-based rejection sampling scheme \cite{razavi2019generating}. Given the conditioning class, we generate multiple image candidates and keep only the one with the highest classifier score for the class. For the external classifier we use a ResNet \cite{he2016deep} with 50 layers. We experiment with acceptance rates of 10\% and 20\% (meaning we keep one out of 10 and one out of 5 images with the highest scores). Results are summarized in Table~\ref{tab:classifier_rejection} and the FID vs. Inception Score curves are plotted in Figure~\ref{fig:classifier_rejection}. We include  a comparison with concurrent work StyleGAN-XL \cite{sauer2022stylegan}. Whilst StyleGAN-XL achieves better FID score, Token-Critic is superior with respect to Inception Score, Precision and Recall.

\subsubsection{Qualitative Results}

\begin{figure}
\centering
\begin{tabular}{c}
Baseline \cite{chang2022maskgit} + Token-Critic (FID 6.80). \\
\includegraphics[width=0.995\textwidth]{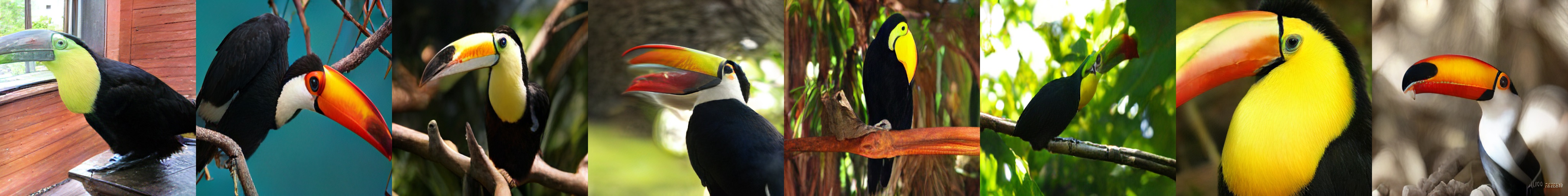}  \\[-0.4em]
\includegraphics[width=0.995\textwidth]{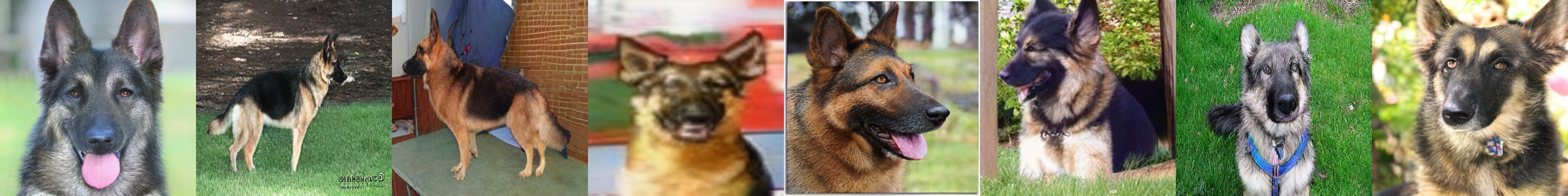} \\[-0.4em]
\includegraphics[width=0.995\textwidth]{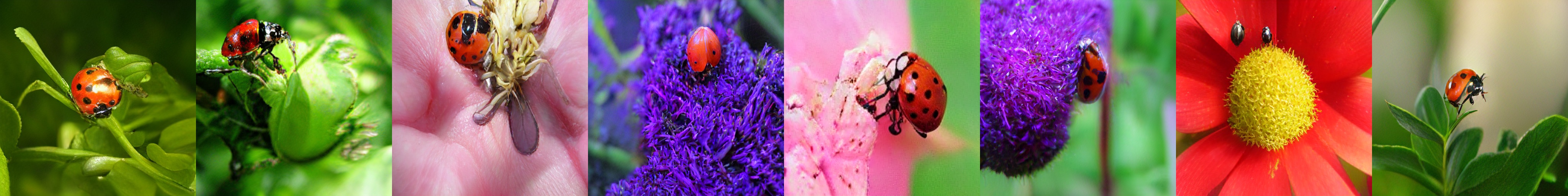}\\[-0.4em]
\includegraphics[width=0.995\textwidth]{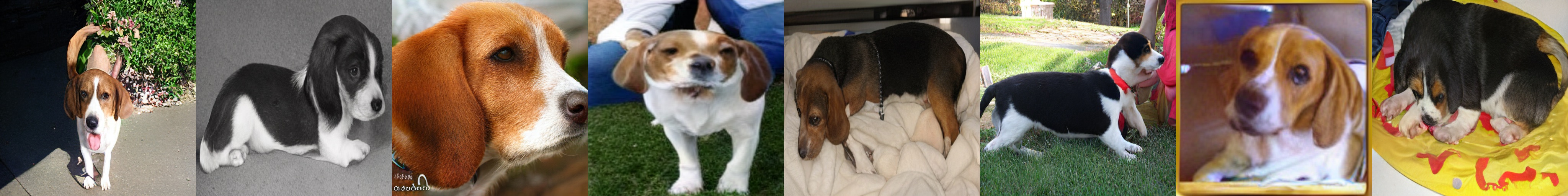}\\[-0.4em]
\includegraphics[width=0.995\textwidth]{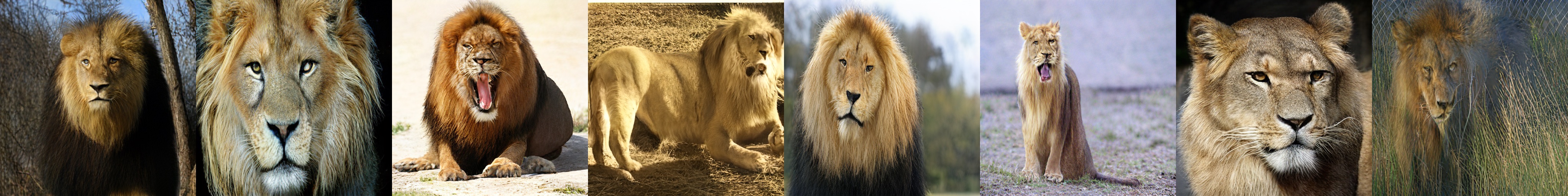}\\
Baseline \cite{chang2022maskgit} (FID 8.48).\\
\includegraphics[width=0.995\textwidth]{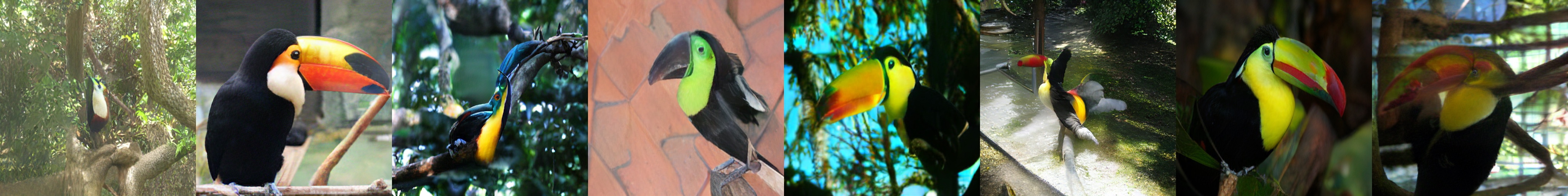}  \\[-0.4em]
\includegraphics[width=0.995\textwidth]{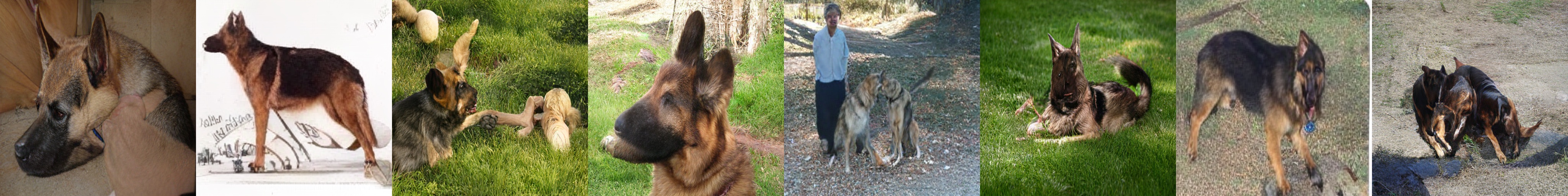} \\[-0.4em]
\includegraphics[width=0.995\textwidth]{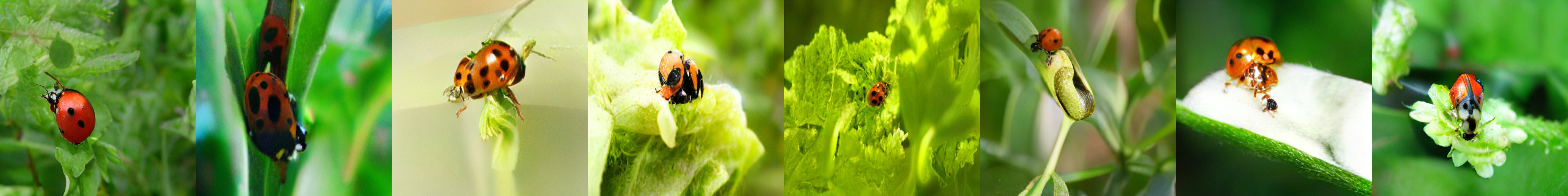}\\[-0.4em]
\includegraphics[width=0.995\textwidth]{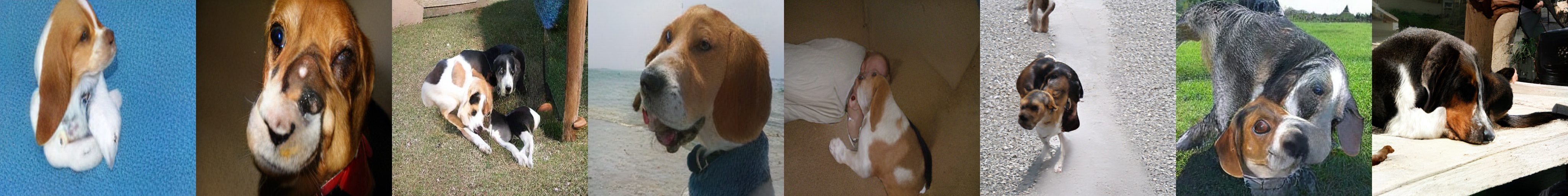}\\[-0.4em]
\includegraphics[width=0.995\textwidth]{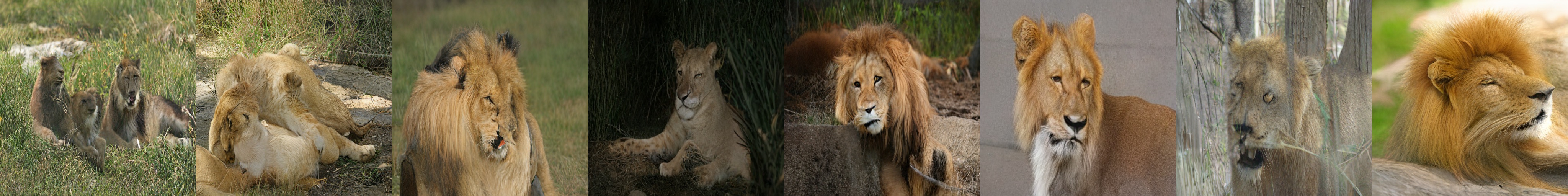}\\
\end{tabular}
\caption{Samples from ImageNet 512x512 models for classes Toucan (96), German Shepherd (235), Ladybug (301), Beagle (162) and Lion (291). All models ran for 18 steps.}
\label{fig:vanilla_examples}
\end{figure}

Figure~\ref{fig:vanilla_examples} shows a qualitative comparison on ImageNet class-conditional generation between the baseline MaskGIT's original sampling procedure \cite{chang2022maskgit} and the proposed sampling using Token-Critic. We demonstrate the models without classifier-based rejection  to better isolate the difference in image quality and diversity obtained by the proposed approach. Notably, sampling with the proposed approach achieves better structural consistency, showing  the ability of the Token-Critic  to capture long range dependencies. We refer to the Appendix for further results and comparisons.

\begin{figure}
\centering
\begin{tabular}{c@{}c@{}c@{}c@{}c@{}c}
\includegraphics[width=0.167\textwidth]{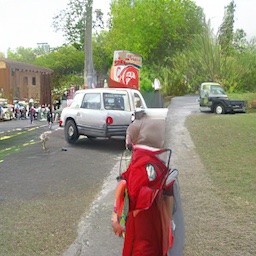}  &
\includegraphics[width=0.167\textwidth]{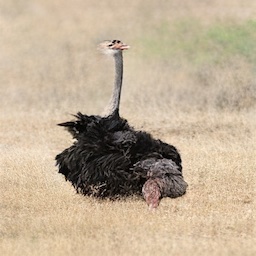}  &
\includegraphics[width=0.167\textwidth]{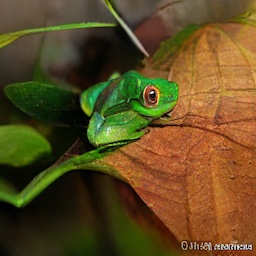}  &
\includegraphics[width=0.167\textwidth]{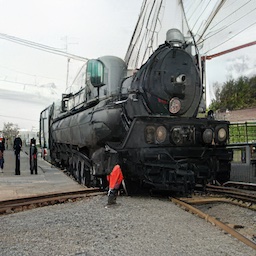}  &
\includegraphics[width=0.167\textwidth]{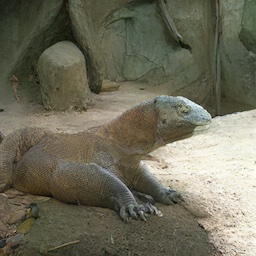}  &
\includegraphics[width=0.167\textwidth]{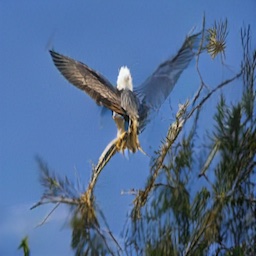}  \\
\includegraphics[width=0.167\textwidth]{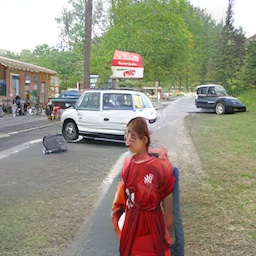}  &
\includegraphics[width=0.167\textwidth]{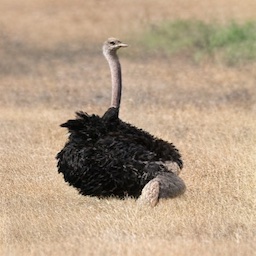}  &
\includegraphics[width=0.167\textwidth]{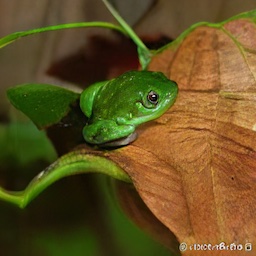}  &
\includegraphics[width=0.167\textwidth]{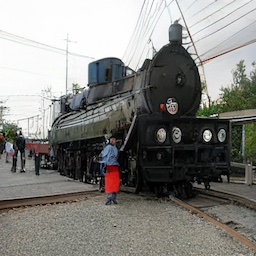}  &
\includegraphics[width=0.167\textwidth]{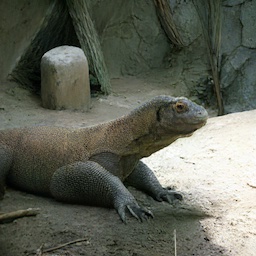}  &
\includegraphics[width=0.167\textwidth]{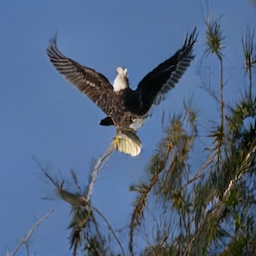}  \\
\end{tabular}
\caption{Refinement of previously generated vector-quantized images by \cite{chang2022maskgit} with an ImageNet 512x512 model. \textbf{Top:} Original samples (FID/IS 8.48/167). \textbf{Bottom:} After refining 60\% of the tokens with lowest Token-Critic score (FID/IS 7.64/182.4). The semantics of the original image is maintained, but it is given a more realistic aspect. }
\label{fig:refinement_examples}
\end{figure}

\subsection{VQ Image Refinement}
To demonstrate the ability of Token-Critic to identify unlikely visual tokens, we apply it to refine the output of the baseline model. Given a generated VQ-encoded image by a MaskGIT generator, we compute the Token-Critic scores on the generated tokens, and proceed to resample the tokens that have low scores. We start by rejecting and replacing 60\% of the original tokens in the first step, and then progressively reject and replace fewer tokens, following again a cosine schedule for $T=9$ steps. The result of this procedure can be regarded as a visual quality improvement of the original images, see Figure~\ref{fig:refinement_examples}. By applying this refinement procedure we improve the FID score of the baseline generator form 6.56 to 5.73 in 256$\times$256 and from 8.48 to 7.64 in 512$\times$512.


\section{Related Work}
While there exist other types of generative models such as VAEs~\cite{kingma2013auto,vahdat2020nvae} and Flow-based models~\cite{rezende2015variational}, we briefly review the works closely relevant to ours.

\noindent\textit{Generative Adversarial Networks (GANs)} are capable of synthesizing high-fidelity images at blazing speeds.
GAN based methods demonstrate impressive capability in yielding high-fidelity samples
\cite{goodfellow2014generative,brock2018large,Karras2019stylegan2}. They suffer from, however, well known issues including training instability and mode collapse which causes a lack of sample diversity. Addressing these issues still remains an active research problem. Note that MaskGIT and Token-Critic are not affected by  adversarial training instability, as the Token-Critic is trained asynchronously over a pre-trained MaskGIT.

\noindent\textit{Generative Image Transformers}
Inspired by the success of the transformer in the NLP field \cite{devlin2018bert,gpt3}, vision transformers~\cite{dosovitskiy2021vit} have been applied to various vision tasks. In particular, the generative image transformer (GIT)~\cite{chen2020imagegpt} is inspired by the generative pre-trained transformer or GPT~\cite{gpt3}. Generally, modern GITs consist of two stages~\cite{Ramesh21dalle}: image quantization and autoregressive decoding, where the former is to compress an image into tokens of a reasonable length whereas the latter, borrowed from neural machine translation~\cite{vaswani2017attention}, generate image tokens as if they were ``visual words''. Most recent contributions are on improving the first stage, \eg, using vector-quantized models of various architectures and losses~\cite{esser2021taming,Ramesh21dalle,vim2021}. Very recently, \cite{zhang2021m6,chang2022maskgit} proposed to use bi-directional transformers to synthesize images, which significantly accelerate the decoding time. Our work builds upon the MaskGIT model~\cite{chang2022maskgit} and improves its mask sampling in non-autoregressive decoding.


\noindent\textit{Denoising Diffusion Models~\cite{sohl2015deep}} define a parameterized Markov chain trained to reverse a forward process of corrupting a training image into pure noise. While many works have focused on continuous (Gaussian) diffusion processes~\cite{kingma2021variational}, closely related to ours are diffusion models with \emph{discrete} state spaces~\cite{seff2019discrete}. For example, Austin~\etal~\cite{austin2021structured} proposed a discrete diffusion (D3) model corrupting data by transition matrices that embed structure knowledge.
Song~\etal~\cite{song2020denoising} introduced implicit diffusion models of non-Markovian diffusion processes. Hoogeboom~\etal~\cite{hoogeboom2021argmax} modeled the categorical data through a fixed multinomial diffusion for image segmentation, which was improved in ImageBART~\cite{esser2021imagebart} by combing with the autoregressive formulation.


The majority of diffusion models is characterized by a forward process with tractable known expressions according to~\cite{austin2021structured}, which is essential for permitting not only efficient forward sampling but also computation of the posterior. From this perspective, our method, similar to MaskGIT~\cite{chang2022maskgit}, is not a traditional diffusion model because it parameterizes the forward process by a transformer that does not have a tractable expression. Since the direct computation of the forward process is intractable in our case, we resort to learning a non-Markov transformer that can teleport to any forward state. This is achieved by the proposed second transformer (Token-Critic).
Empirically, we found this strategy to be effective needing considerably fewer number of decoding steps (typically 8-16 steps) while producing competitive quality.


\section{Conclusion}
In this work, we proposed a novel method for sampling from a non-autoregressive generative vision transformer. It is based on using a second transformer, the Token-Critic, to select which tokens are accepted and which are rejected and resampled during the iterative generative process. Given a reconstructed masked image, the Token-Critic is trained to distinguish which visual tokens  belong to the original image and which are predictions of the generative transformer. Coupled with the Token-Critic, an already powerful non-autoregressive transformer significantly improves its performance, and outperforms the state-of-the-art in terms of the trade-off between generated image quality and variety, in the challenging task of class-conditional ImageNet generation.

%
%
\bibliographystyle{splncs04}
\bibliography{main}

\clearpage

\appendix
\section{Comparison to related work on ImageNet 512x512}
\subsection{Base models }

\begin{figure}
\centering
\begin{tabular}{c}
\includegraphics[width=0.985\textwidth,trim=0 0 0 0,clip]{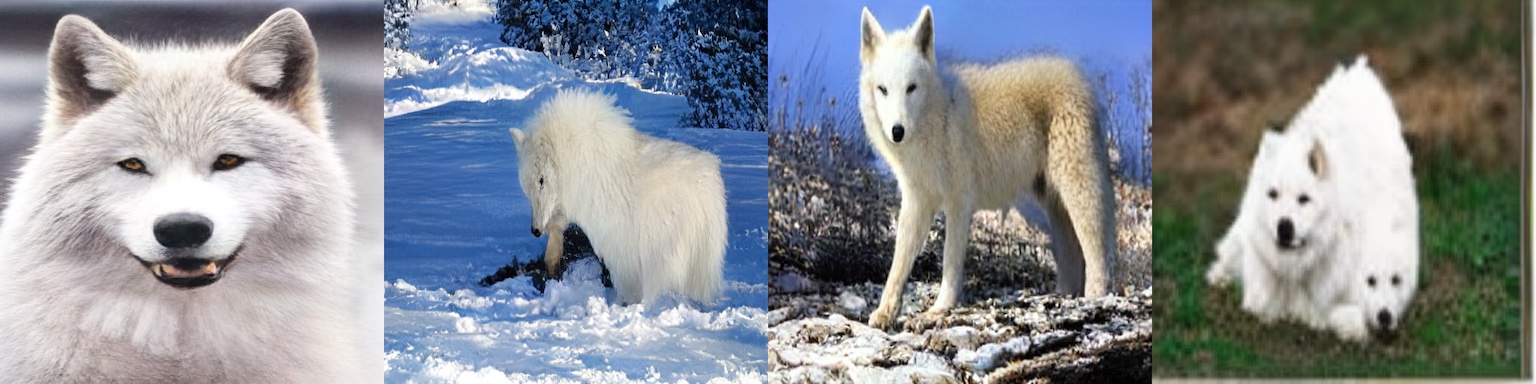}  \\[-.36em]
\includegraphics[width=0.985\textwidth,trim=0 0 0 0,clip]{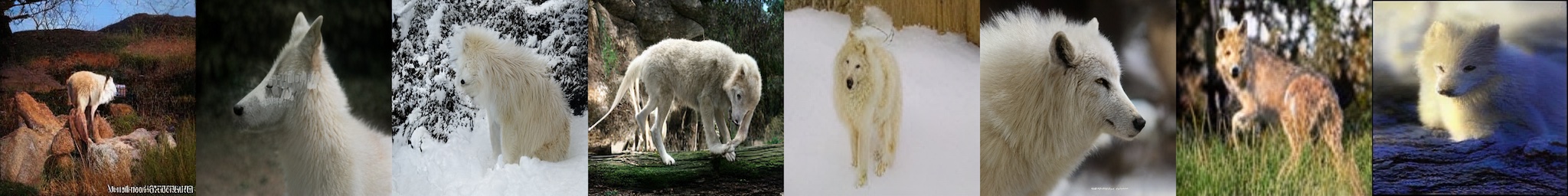}  \\
(a) Token-Critic (FID/IS 6.80/182.1)\\
\includegraphics[width=0.985\textwidth,trim=0 0 0 0,clip]{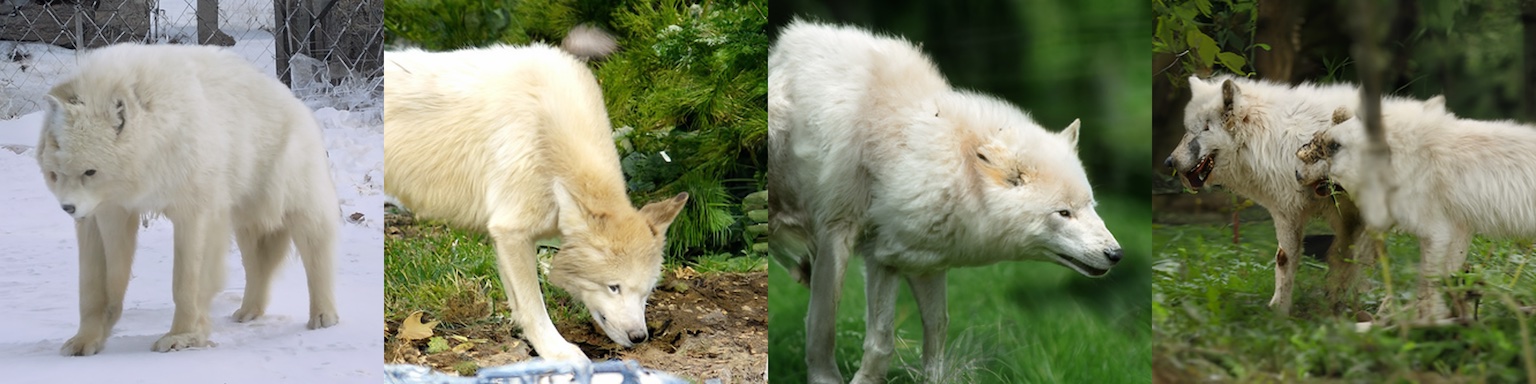}  \\[-.36em]
\includegraphics[width=0.985\textwidth,trim=0 0 0 0,clip]{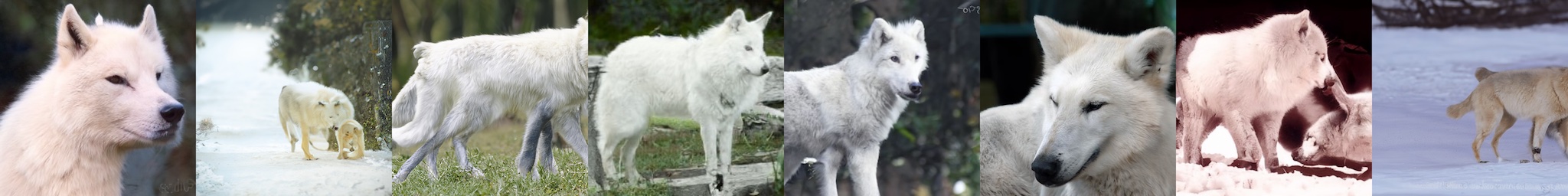}  \\
(b) ADM+G  (FID/IS 7.72/172.7)\\
\includegraphics[width=0.985\textwidth,trim=0 0 0 0,clip]{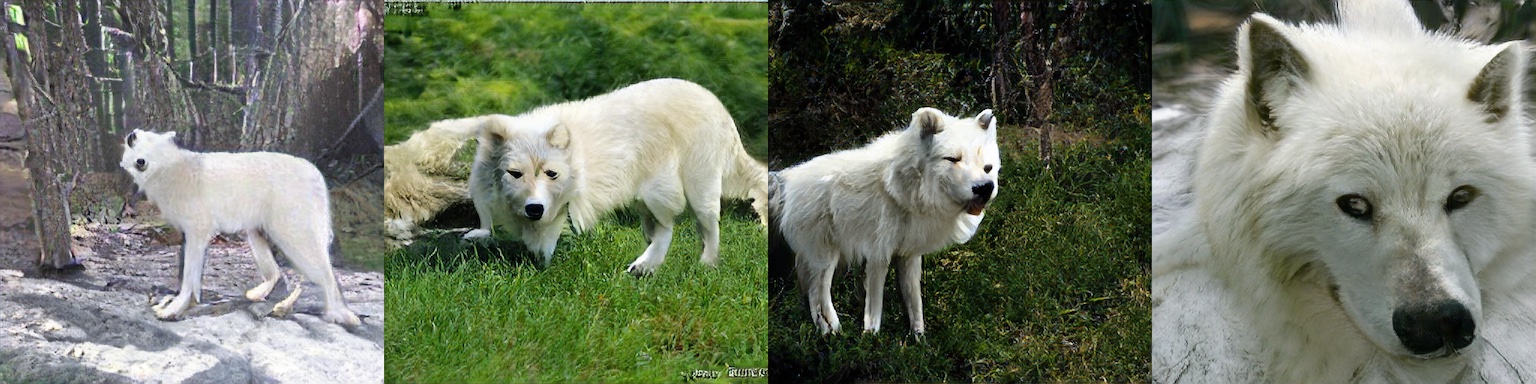}  \\[-.36em]
\includegraphics[width=0.985\textwidth,trim=0 0 0
0,clip]{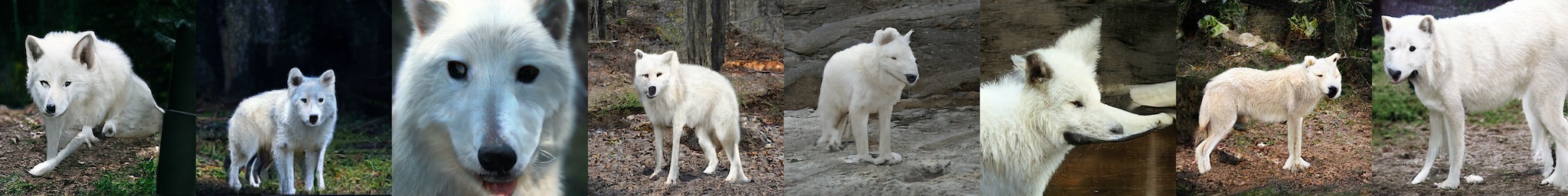}  \\
(c) BigGAN (FID/IS 8.43/177.9)\\
\end{tabular}
\caption{Comparison on 512x512 class-conditional image generation on ImageNet class  ``White wolf'' (270).}
\label{fig:512_2}
\end{figure}

\begin{figure}
\centering
\begin{tabular}{c}
\includegraphics[width=0.985\textwidth,trim=0 0 0 0,clip]{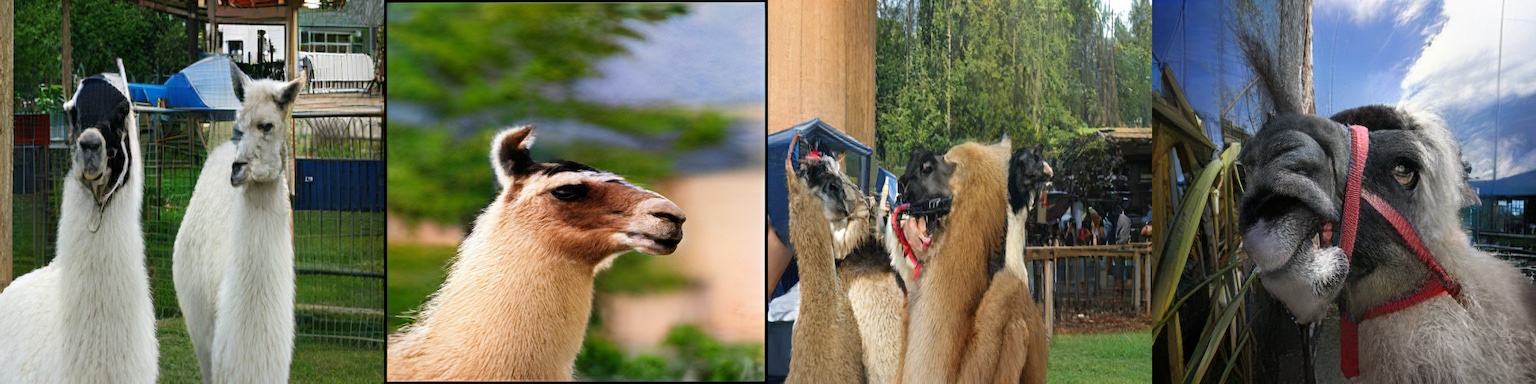}  \\[-.36em]
\includegraphics[width=0.985\textwidth,trim=0 0 0 0,clip]{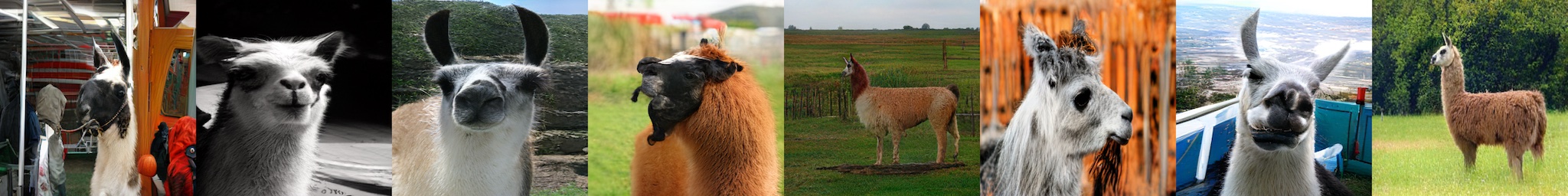}  \\
(a) Token-Critic (FID/IS 6.80/182.1)\\
\includegraphics[width=0.985\textwidth,trim=0 0 0 0,clip]{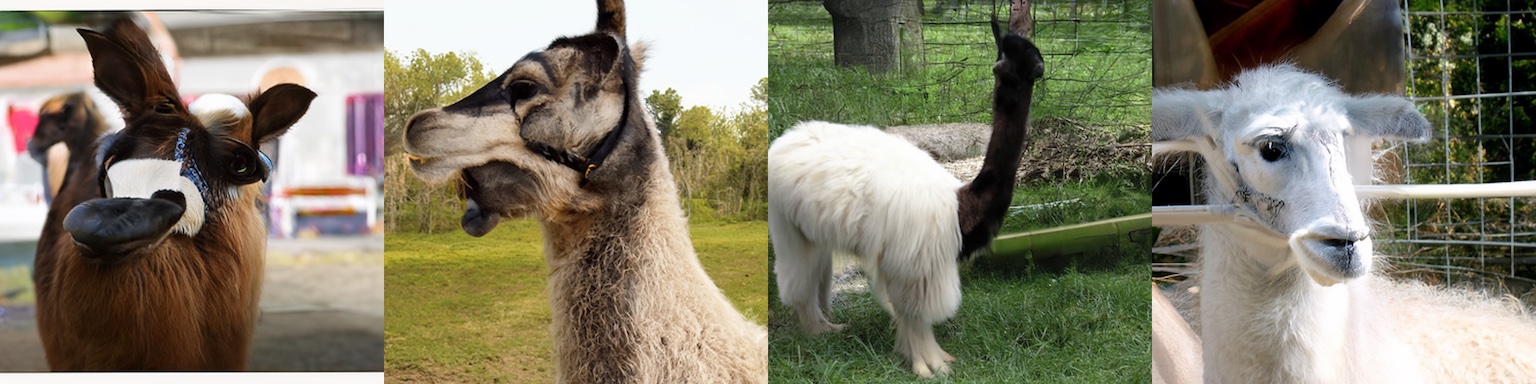}  \\[-.36em]
\includegraphics[width=0.985\textwidth,trim=0 0 0 0,clip]{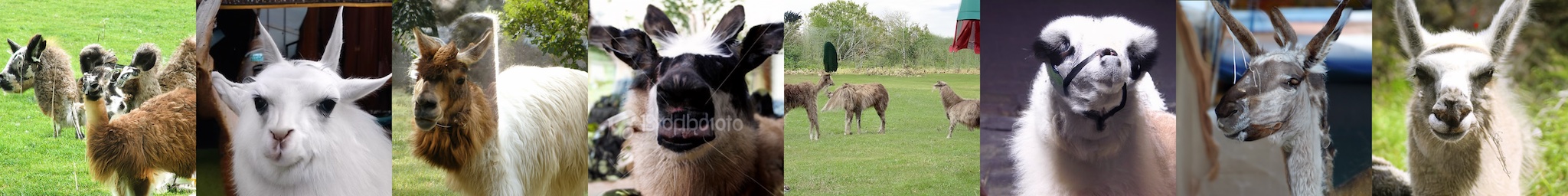}  \\
(b) ADM+G  (FID/IS 7.72/172.7)\\
\includegraphics[width=0.985\textwidth,trim=0 0 0 0,clip]{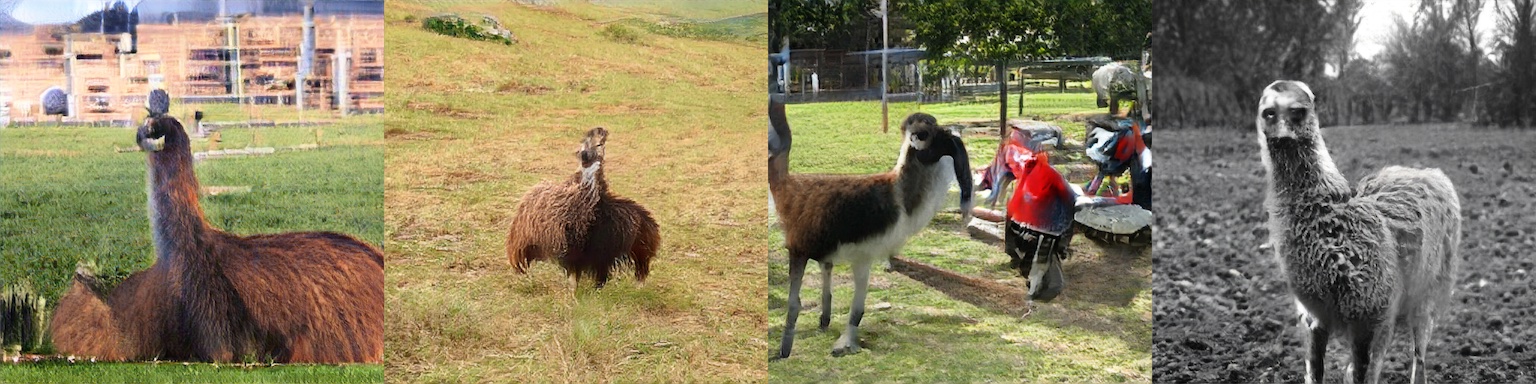}  \\[-.36em]
\includegraphics[width=0.985\textwidth,trim=0 0 0
0,clip]{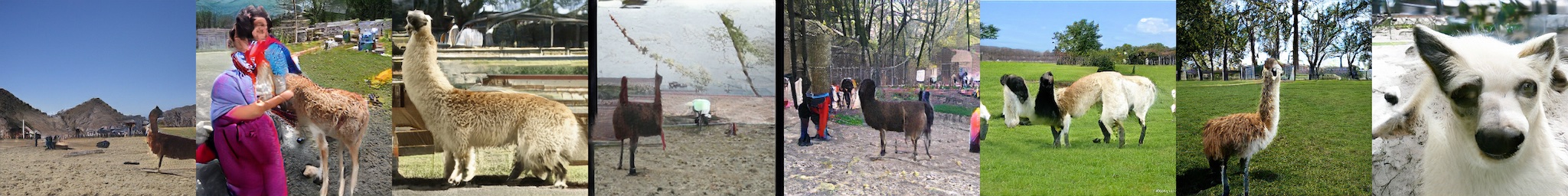}  \\
(c) BigGAN (FID/IS 8.43/177.9)\\

\end{tabular}
\caption{Comparison on 512x512 class-conditional image generation on ImageNet class  ``Llama'' (355).}
\label{fig:512_3}
\end{figure}

\begin{figure}
\centering
\begin{tabular}{c}
\includegraphics[width=0.985\textwidth,trim=0 0 0 0,clip]{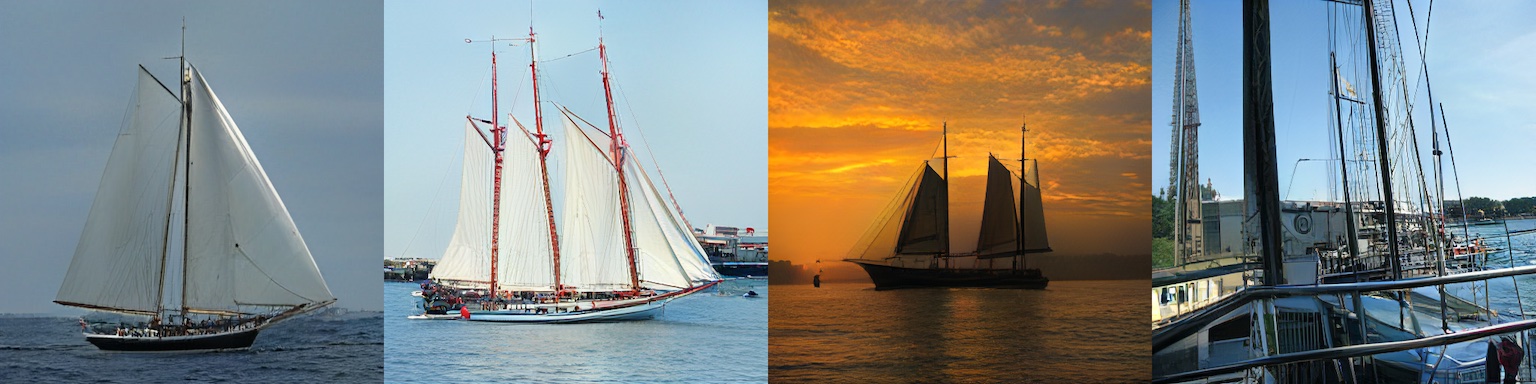}  \\[-.36em]
\includegraphics[width=0.985\textwidth,trim=0 0 0 0,clip]{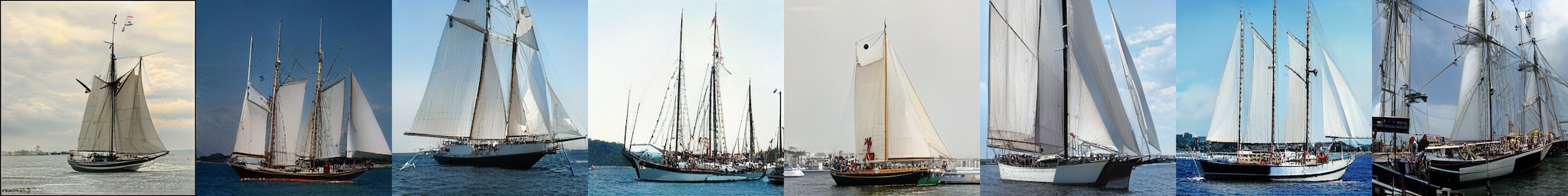}  \\
(a) Token-Critic (FID/IS 6.80/182.1)\\
\includegraphics[width=0.985\textwidth,trim=0 0 0 0,clip]{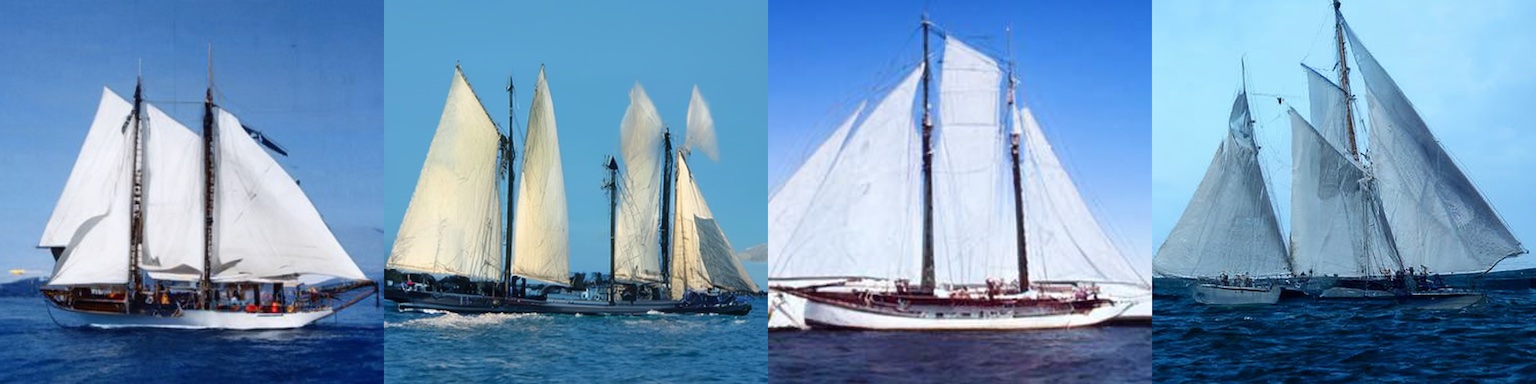}  \\[-.36em]
\includegraphics[width=0.985\textwidth,trim=0 0 0 0,clip]{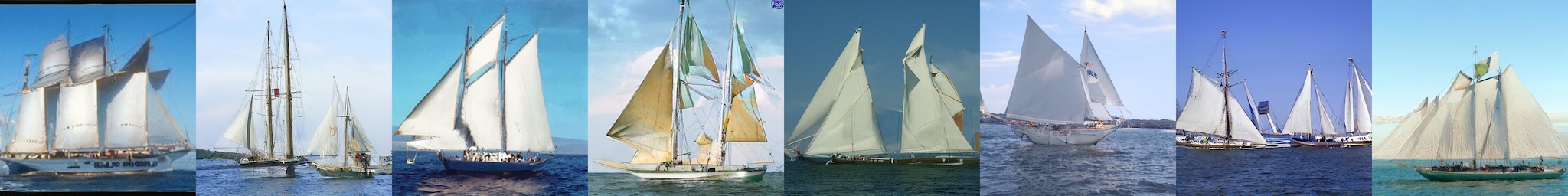}  \\
(b) ADM+G  (FID/IS 7.72/172.7)\\
\includegraphics[width=0.985\textwidth,trim=0 0 0 0,clip]{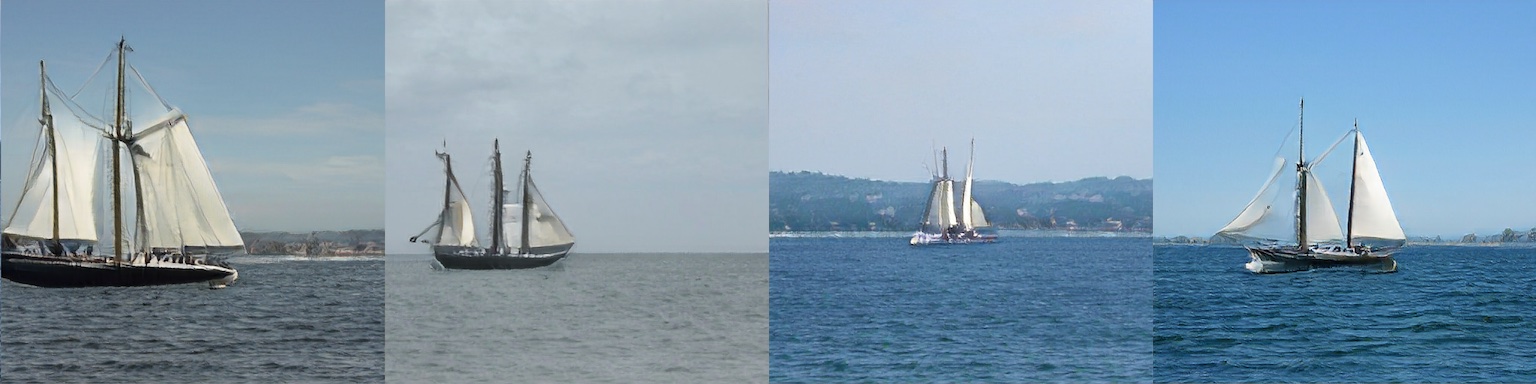}  \\[-.36em]
\includegraphics[width=0.985\textwidth,trim=0 0 0
0,clip]{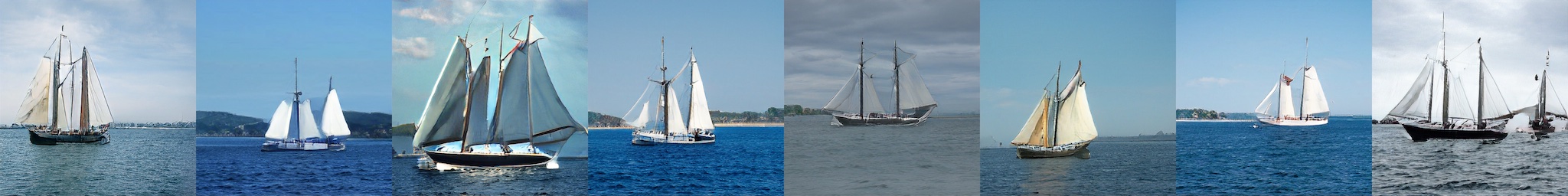}  \\
(c) BigGAN (FID/IS 8.43/177.9)\\
\end{tabular}
\caption{Comparison on 512x512 class-conditional image generation on ImageNet class ``Schooner'' (780).}
\label{fig:512_4}
\end{figure}

In Figures~\ref{fig:512_2},~\ref{fig:512_3} and~\ref{fig:512_4} we compare the result of sampling from Token-Critic with one competing GAN, BigGAN \cite{brock2018large}, and one diffusion model, ADM with classifier guidance (ADM+G)~\cite{dhariwal2021diffusion}. We compare on ImageNet 512x512 as this is the more challenging case. 

Our goal here is to directly compare the performance of the base models in capturing the class-conditional distributions of 512x512 real images. Thus, we do not include classifier rejection for Token-Critic or upsampling for ADM, as the resulting samples would depend on a separate process.

Results for ADM+G \cite{dhariwal2021diffusion} were obtained using the authors' publicly available source code\footnote{\url{https://github.com/openai/guided-diffusion}}. Results for BigGAN \cite{brock2018large} were obtained using the authors' implementation. Note that BigGAN uses one step, ADM+G 1000  steps, and Token-Critic 18 generator steps and 18 critic steps.

\subsection{Combined models }
\begin{figure}
\centering
\begin{tabular}{c}
\includegraphics[width=0.985\textwidth,trim=0 0 0 0,clip]{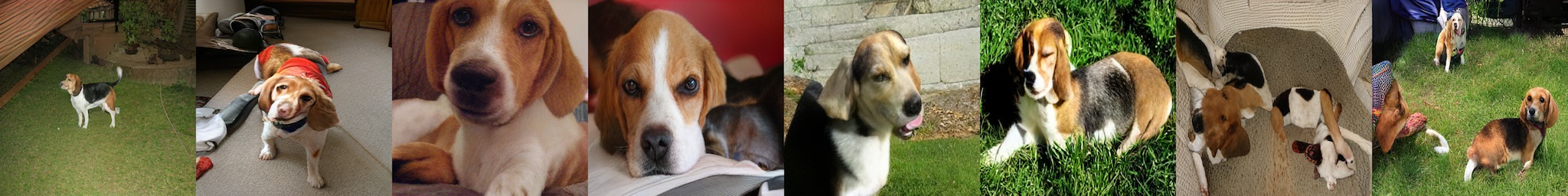}  \\[-.36em]
\includegraphics[width=0.985\textwidth,trim=0 0 0 0,clip]{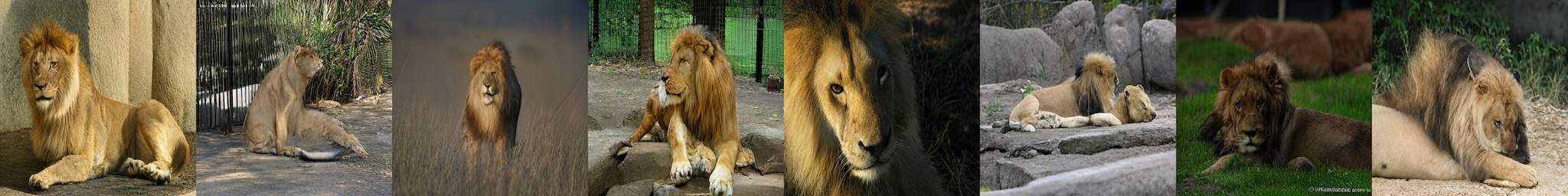}  \\[-.36em]
\includegraphics[width=0.985\textwidth,trim=0 0 0 0,clip]{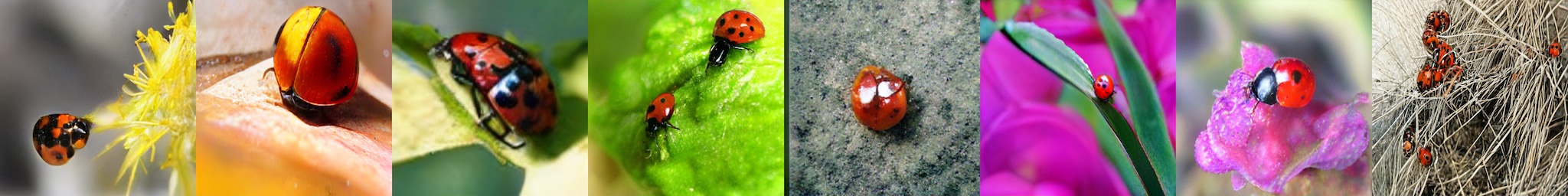}  \\[-.36em]
\includegraphics[width=0.985\textwidth,trim=0 0 0 0,clip]{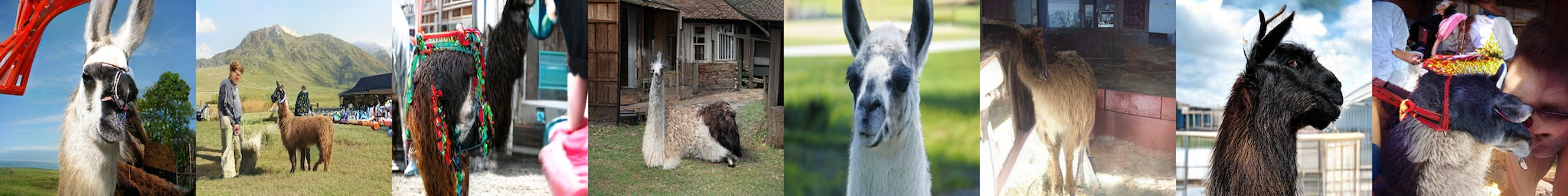}  \\
(a) Token-Critic + Classifier-based rejection (FID/IS 4.03/305.2)\\
\includegraphics[width=0.985\textwidth,trim=0 0 0 0,clip]{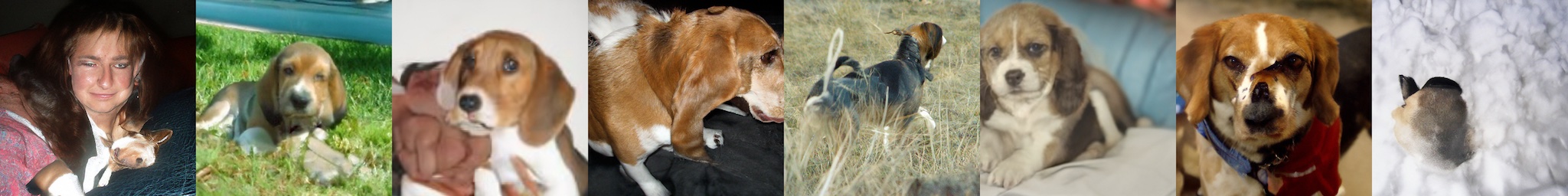}  \\[-.36em]
\includegraphics[width=0.985\textwidth,trim=0 0 0 0,clip]{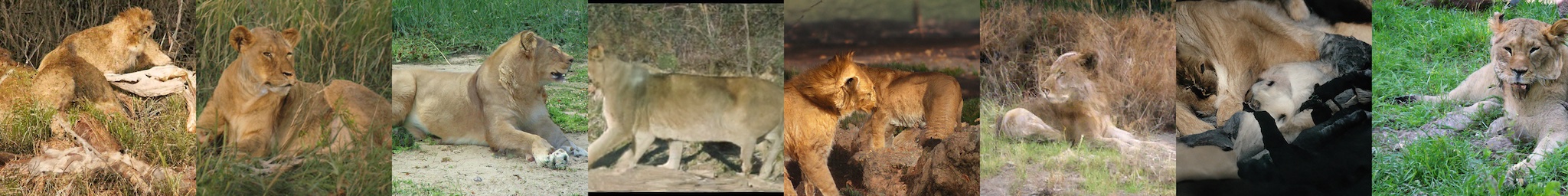}  \\[-.36em]
\includegraphics[width=0.985\textwidth,trim=0 0 0 0,clip]{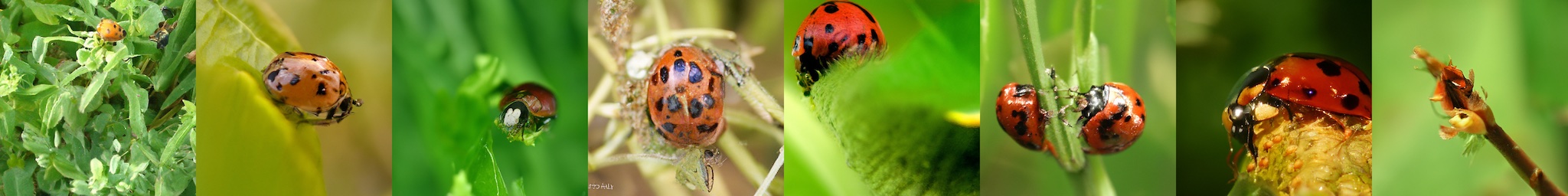}  \\[-.36em]
\includegraphics[width=0.985\textwidth,trim=0 0 0 0,clip]{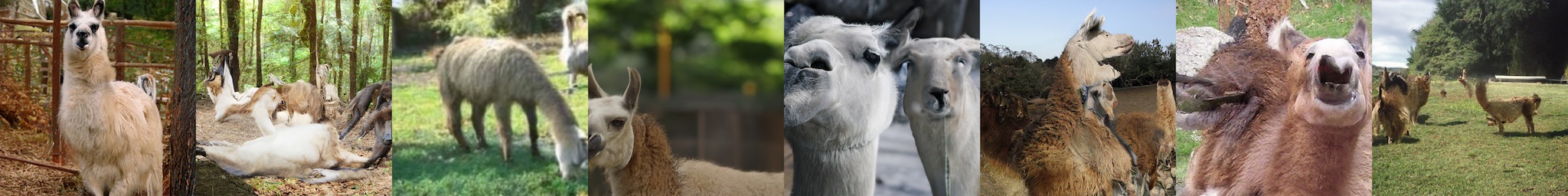}  \\
(b) ADM + Guidance + Upsampling  (FID/IS 3.85/221.7)\\
\end{tabular}
\caption{Comparison on 512x512 class-conditional image generation with ADM+G+U \cite{dhariwal2021diffusion}, for ImageNet classes ``Beagle'' (162), ``Lion'' (291), ``Ladybug'' (301) and ``Llama'' (355).}
\label{fig:external}
\end{figure}

In Figure~\ref{fig:external} we compare the models that obtain better FID and Inception scores in Table~2 by leveraging an external process. For Token-Critic, the external process is classifier-based rejection sampling using a ResNet50 classifier. For ADM with guidance and upsampling (ADM+G+U) \cite{dhariwal2021diffusion}, the external process consists in using an upsampling diffusion model to rescale samples from 128x128 to 512x512. Results for \cite{dhariwal2021diffusion} were obtained using the authors' publicly available source code. Note that ADM+G+U uses 250  steps for 128x128 generation and 250  steps for upsampling. Token-Critic with rejection sampling with 20\% acceptance rate uses five times 18 generator steps and 18 critic steps.

\clearpage
\section{On Token-Critic training objective}

As motivated in the main manuscript, we seek to match the distributions of 1) real masked images  and 2)  masked images obtained by the proposed method, which combines the (fixed) generator $p_\theta(\hat{\bx}_0|\bx_t)$ and the Token-Critic model $p_\phi(\bx_t|\hat{\bx}_0)$: 

\begin{align}
    KL\left(q(\bx_t) \Vert p_{\theta,\phi}(\bx_t)\right) &= - \mathds{E}_{q(\bx_t)} \log \frac{p_{\theta,\phi}(\bx_t)}{q(\bx_t)}\\
     &=  - \mathds{E}_{q(\bx_t)} \log \sum_{\hat{\bx}_0}
     \frac{p_{\phi}(\bx_t | \hat{\bx}_0)p_\theta(\hat{\bx}_0 )}{q(\bx_t)}\\
     &=  - \mathds{E}_{q(\bx_t)} \log \sum_{\hat{\bx}_0:~ \hat{\bx}_0\odot(\mathds{1}-\bm_t)=\bx_t\odot(\mathds{1}-\bm_t)}
     \frac{p_{\phi}(\bx_t | \hat{\bx}_0)p_\theta(\hat{\bx}_0 )}{q(\bx_t)} \label{eq:subsum}\\
&\approx  - \mathds{E}_{q(\bx_t)} \log \sum_{\hat{\bx}_0}
     \frac{p_{\phi}(\bx_t | \hat{\bx}_0)p_\theta(\hat{\bx}_0 | \bx_t )}{q(\bx_t ) Z_\theta(\bx_t)} \label{eq:approx}\\
 &\leq - \mathds{E}_{q(\bx_t)}\mathds{E}_{p_\theta(\hat{\bx}_0 | \bx_t)} \log 
      p_{\phi}(\bx_t | \hat{\bx}_0) + C.       \label{eq:jensen}
    %
    %
    %
\end{align}
In  \eqref{eq:subsum} we used the fact that $p_\phi(\bx_t|\hat{\bx}_0)$ is $0$ whenever the unmasked elements of $\bx_t$ and $\hat{\bx}_0$ do not match, since the critic only determines where to mask. In  \eqref{eq:approx} we assume a perfect model $p_\theta$ and (noting that the generator $G_\theta(\bx_t)\rightarrow \hat{\bx}_0$ always copies the unmasked tokens in the output),  for the $\hat{\bx}_0$ that are compatible with $\bx_t$, we replace $p_\theta(\hat{\bx}_0)$ with the conditional $p_\theta(\hat{\bx}_0|\bx_t)$ and a renormalizing factor $Z_\theta(\bx_t)$. In \eqref{eq:jensen} we used Jensen's inequality, and use $C$ for the terms that do not depend on $\phi$ (recall that $\theta$ is frozen). 

Therefore, under the assumptions, the  objective in \eqref{eq:jensen}  upper bounds the divergence between the the real and generated masked images. Finally, since $\bx_t$ is  completely determined by $\hat{\bx}_0$ and the mask $\bm_t$, we train $p_\phi$ to maximize the  likelihood of the sampled mask $\bm_t$ (Equation \ref{eq:token-critic_train}).

\end{document}